%% file: main.tex
\documentclass[10pt,twocolumn,letterpaper]{article}

\usepackage{iccv}
\usepackage{times}
\usepackage{epsfig}
\usepackage{graphicx}
\usepackage{amsmath}
\usepackage{amssymb}
\usepackage{times}
\usepackage{epsfig}
\usepackage{graphicx}
\usepackage{amsmath}
\usepackage{amssymb}
\usepackage{cellspace}
\usepackage{xcolor}
\usepackage{float}
\restylefloat{table}
\usepackage{diagbox}
\usepackage{wrapfig}
\usepackage{subfig}
\usepackage{multirow}
\usepackage{mathtools}
\usepackage{booktabs}
\usepackage{kantlipsum}
\usepackage{gensymb}
\usepackage{longtable,rotating}
\usepackage{threeparttable}
\usepackage[accsupp]{axessibility}
\usepackage{pifont}
\usepackage{nicefrac}
\usepackage{paralist}
\usepackage{algorithm}
\usepackage[noend]{algpseudocode}
\newcommand{\cmark}{\text{\ding{51}}}
\newcommand{\xmark}{\text{\ding{55}}}

\newcommand*{\affmark}[1][*]{\textsuperscript{#1}}

\usepackage[breaklinks=true,bookmarks=false]{hyperref}

\iccvfinalcopy 

\def\iccvPaperID{7007} 

\ificcvfinal\fi

\begin{document}

\title{MEDIRL: Predicting the Visual Attention of Drivers via \\
Maximum Entropy Deep Inverse Reinforcement Learning}

\input{sections/author}
\maketitle
\ificcvfinal\thispagestyle{empty}\fi

\begin{abstract}
\input{sections/abstract}

\end{abstract}
\section{Introduction}
\input{sections/introduction}
\section{Related Work}
\input{sections/related-works}
\section{Method}\label{sec:medirl}
\input{sections/method}

\section{The EyeCar Dataset}\label{sec:eyecar}
\input{sections/eyecar}
\section{Experiments}\label{sec:experiment}
\input{sections/experiment}
\section{Results}\label{sec:results}
\input{sections/result}
\section{Conclusion}
\input{sections/conclusion}

\input{sections/acknowledgment}

\input{sections/supplementary}


{\small
\bibliographystyle{ieee_fullname}
\bibliography{egbib}
}

\end{document}

%% file: sections/author.tex
\author{Sonia Baee\affmark[1], Erfan Pakdamanian\affmark[1], Inki Kim\affmark[2], Lu Feng\affmark[1], Vicente Ordonez\affmark[3], Laura Barnes\affmark[1]\\
\affmark[1]University of Virginia, \affmark[2]University of Illinois at Urbana Champaign, \affmark[3]Rice University\\
{\tt\small sb5ce@virginia.edu, ep2ca@virginia.edu, inkikim@illinois.edu}\\
{\tt\small Lu.feng@virginia.edu, vicenteor@rice.edu, lb3dp@virginia.edu}
}

%% file: sections/abstract.tex
Inspired by human visual attention, we propose a novel inverse reinforcement learning formulation using Maximum Entropy Deep Inverse Reinforcement Learning (MEDIRL) for predicting the visual attention of drivers in accident-prone situations. MEDIRL predicts fixation locations that lead to maximal rewards by learning a task-sensitive reward function from eye fixation patterns recorded from attentive drivers. Additionally, we introduce EyeCar, a new driver attention dataset in accident-prone situations. We conduct comprehensive experiments to evaluate our proposed model on three common benchmarks: (DR(eye)VE, BDD-A, DADA-2000), and our EyeCar dataset. Results indicate that MEDIRL outperforms existing models for predicting attention and achieves state-of-the-art performance. We present extensive ablation studies to provide more insights into different features of our proposed model.\footnote{The code and dataset are provided for reproducibility in \url{https://github.com/soniabaee/MEDIRL-EyeCar}.
}




%% file: sections/introduction.tex
Autonomous vehicles have witnessed significant advances in recent years. These vehicles promise better safety and freedom from the prolonged and monotonous task of driving. However, one of the remaining safety challenges of vision-based models integrated into these vehicles is how to quickly identify important visual cues and understand risks involved in traffic environments at a time of urgency~\cite{tawari2017computational}. Humans have an incredible visual attention ability to quickly detect the most relevant stimuli, to direct attention to potential hazards in complex situations~\cite{pakdamanian2021deeptake}, and to select only a relevant fraction of perceived information for more in-depth processing~\cite{ungerleider2000mechanisms}. Humans are able to guide their attention by a combination of bottom-up~(\textit{stimuli driven}, e.g., color and intensity) and top-down~(\textit{task driven}, e.g., current goals or intention) mechanisms~\cite{deng2016does,katsuki2014bottom}.


\begin{figure}[t]
    \centering
     \includegraphics[width=0.98\linewidth]{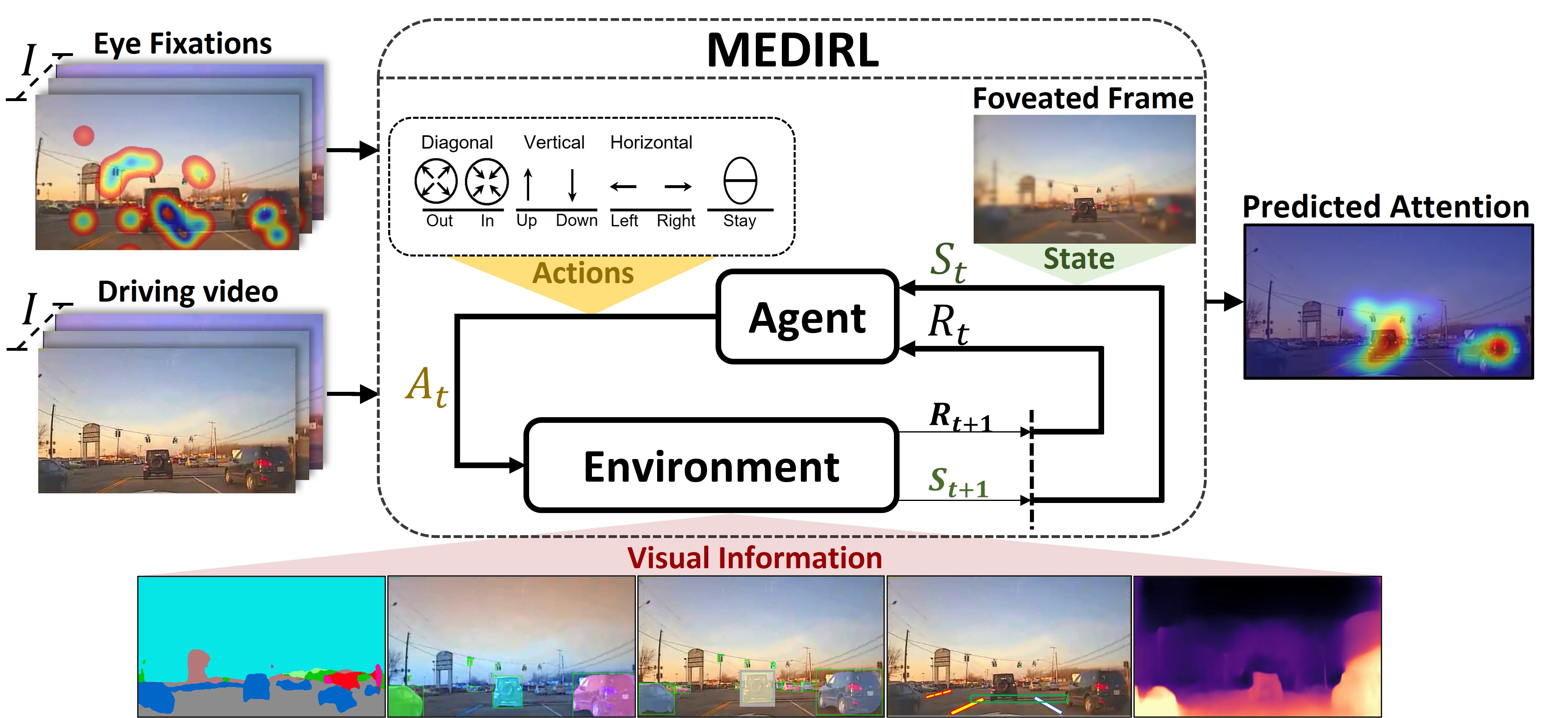}
     \vspace{-0.5mm}
    \caption{Given a driving video and corresponding eye fixations as inputs, MEDIRL learns to model the fixation selection as a sequence of states and actions~$(S_t, A_t)$. MEDIRL then predicts a maximally-rewarding fixation location by perceptually parsing a scene to extract rich visual information~(environment) and accumulating a sequence of visual cues through fixations~(state).
    }
    \label{fig:problem-setup}
    \vspace{-0.15in}
\end{figure}

During \emph{task-specific} activities, the \emph{goal-directed} behavior of humans along with their underlying \emph{target-based} selective attention, enables drivers to ignore objects and unnecessary details in their field of view that are irrelevant to their decisions~\cite{chen2012goal,chen2015deepdriving}. For example, at one moment, a driver's goal might be to initiate an overtaking maneuver, thus a nearby vehicle becomes the target object. Later, the driver may need to stop abruptly to avoid an accident, thereby the brake light of the car in front becomes the target object. Despite recent progress in computer vision models for autonomous systems~\cite{kim2017interpretable,xu2017end}, they are still behind the foveal vision ability of humans~\cite{ohn2020learning,xia2020periphery,zelinsky2019benchmarking}. 

Inverse reinforcement learning~(IRL) algorithms are capable to address this problem by learning to imitate the efficient attention allocation produced by an expert,~i.e.,~an attentive driver~\cite{ng2000algorithms}. It is important that autonomous vehicles leverage human visual attention mechanisms to improve their performance, especially for better safety in critical situations where rare events can be encountered. In this paper, we introduce Maximum Entropy Deep Inverse Reinforcement Learning~(MEDIRL) to learn 
\emph{task-specific} visual attention policies to reliably predict attention in imminent rear-end collisions.

Prior efforts in bottom-up saliency models commonly prioritize pixel location~(e.g., free-viewing fixation)~\cite{kruthiventi2017deepfix,pal2020looking,stojic2020uncertainty}. These models do not fully capture driver attention in goal-directed behavior~\cite{einhauser2020fixation, xia2020periphery,xia2020periphery,kummerer2017understanding}. Moreover, video-based saliency models usually aggregate spatial features guided by saliency maps in each frame~\cite{wang2018revisiting,jiang2017predicting,hu2020dgaze,yang2019video}. However, most of these fixation prediction models utilized a particular source of information~\cite{xia2020periphery,palazzi2018predicting,fang2019dada}, and did not consider to jointly process spatial and temporal information~\cite{wang2018revisiting,hu2020dgaze}. In this work, we aim to predict eye fixation patterns made prior to critical situations, where these patterns can be either spatial~(fixation map) or spatiotemporal~(fixation sequences) features.

Inverse reinforcement learning (IRL) is an advanced form of imitation learning~\cite{ziebart2008maximum,wulfmeier2015maximum} that enables a learning agent to acquire skills from expert demonstrations~\cite{tschiatschek2019learner}. Our proposed MEDIRL model learns a sequence of eye fixations by considering each fixation as a potential source of reward~\cite{yang2020predicting}. We leverage collective visual information that has been deemed relevant for video saliency in prior works~\cite{min2019tased,pal2020looking,chen2021video}. For example, if an autonomous system tries to locate the salient regions of a driving scene before an imminent rear-end collision, the desired visual behavior can be demonstrated by studying the attention of a driver who effectively detects brake lights. In this way, the learning agent can infer a reward function explaining experts' behavior and optimize its own behavior accordingly. To this end, our proposed model predicts driver attention where a fixation pattern is represented as state-action pairs. Given a video frame input paired with eye fixations, MEDIRL predicts a maximally-rewarding fixation location~(action) by perceptually parsing a scene to extract rich visual information~(environment), and accumulating a sequence of visual cues through fixations~(state)~(see Figure~\ref{fig:problem-setup}).

Additionally, we introduce \textit{EyeCar}, a new driver attention dataset in accident-prone situations. EyeCar is essential for training goal-directed attention models as it is the only dataset capturing attention before accidents in an environment with high traffic density.
We exhaustively evaluate our proposed model on three common benchmarks~(DR(eye)VE~\cite{palazzi2018predicting}, BDD-A~\cite{xia2018predicting}, DADA-2000~\cite{fang2019dada}) as well as our own EyeCar dataset. The experimental results show that MEDIRL outperforms state-of-the-art models on driver attention prediction. We also conduct extensive ablation studies to determine which input features are most important for driver attention prediction in critical situations.

Our \textbf{contributions} can be summarized as follows:
\begin{compactitem} 
    \item We propose MEDIRL, a novel IRL formulation for predicting driver visual attention in accident-prone situations. MEDIRL uses maximum entropy deep inverse reinforcement learning to predict maximally-rewarding fixation locations.
    \item  We introduce EyeCar, a new driver attention dataset comprised of  rear-end collisions videos for the goal-directed attention problem in critical driving situations.
    \item Extensive experimental evaluation on three driver attention benchmark datasets:~DR(eye)VE~\cite{palazzi2018predicting}, BDD-A~\cite{xia2018predicting}, DADA-2000~\cite{fang2019dada}, and EyeCar. Results show that MEDIRL outperforms existing models for attention prediction and achieves state-of-the-art performance. Besides, we present ablation studies showing  target~(brake light), non-target~(context), and driving tasks are important for predicting driver attention.
\end{compactitem}

%% file: sections/related-works.tex
Our work is broadly related to prior efforts on models for fixation prediction, using inverse reinforcement learning for visual tasks, and prior datasets for driving tasks.

\noindent{\textbf{Fixation Prediction.}} With increased access to large-scale annotated attention datasets and advanced data-driven machine learning techniques, prediction of human saliency has received significant interest in computer vision~\cite{wang2019learning, wang2017deep, kruthiventi2017deepfix, zhong2013video, cornia2018predicting, min2019tased}. A large number of previous studies explored bottom-up saliency models and visual search strategies over static stimuli~\cite{fan2019shifting, li2015visual, gong2015saliency, fu2015normalized,borji2015cat2000,yun2013studying}, and video~\cite{zhong2013video, wang2015consistent, mathe2014actions,min2019tased,zahedian2019localization,chen2021video}. Generally, the output of these models is an attention map showing the probability of eye fixation distribution. 
In contrast to this approach, fewer works explored top-down attention models for explaining sequences of eye movements~\cite{ sprague2004eye,borji2012probabilistic, borji2010online}. 
More recently, some works explored visual attention models in the context of driving~\cite{guangyu2019dbus,xia2020periphery, gao2019goal}. 
Because task-specific instructions may change gaze distributions~\cite{rothkopf2007task}, some models commonly detect salient regions of images or videos in a free-viewing task. Prior research also studied the pattern of eye movements associated with the task-specific activities~\cite{mathe2014actions, anderson2018bottom}. Some of these works rely on the direct ties between eye movement and the demands of a task~\cite{yang2020predicting,tatler2011eye,sprague2004eye}. These previously proposed attention models are trained mostly on static image-viewing scenarios while human attention typically gets information in a sequential fashion.
Further, recent video-saliency works have proposed joint bottom-up and top-down mechanisms for attention prediction using deep learning~\cite{palazzi2018predicting, xia2018predicting,fang2019dada, kim2020advisable,pal2020looking}. However, they did not consider to jointly process spatial and temporal information. 
We are interested in detecting the salient regions of a scene in a task-specific driving activity in which \textit{estimating where the drivers are dynamically looking at}, and \textit{reliably detecting the task-related objects~(target objects)}. 

\noindent{\textbf{Inverse Reinforcement learning.}}~Our approach builds on works on modeling human visual attention with their fixation being a sequential decision process of the agent to detect salient regions~\cite{ mathe2013action,zelinsky2020predicting,liu2019gaze}. The recently proposed work by Yang \textit{et al.}~\cite{yang2020predicting} is the closest to our work as it proposes a model of visual attention in a visual search task of~\textit{static images}. We go further by addressing video saliency predictions in a dynamic and complex driving environment. Our model also does not require to predefine a set of targets but instead parses each driving video frame to extract rich scene context and candidate target objects. Next, it integrates visual cues with driver's eye fixations. It then recovers the intrinsic task-specific reward functions~\cite{zheng2018learning} induced by visual attention allocation policies recorded from drivers in a driving environment. To do that, we propose to use maximum entropy deep IRL~\cite{ziebart2008maximum} which can handle raw image inputs and enables the model to handle the often sub-optimal and seemingly stochastic behaviors of drivers~\cite{wulfmeier2015maximum}.

\noindent{\textbf{Driving Attention Datasets.}}~Several driving behavior datasets have been proposed~\cite{codevilla2019exploring,xu2017end, Ramanishka_2018_CVPR}. However, only a few large-scale, publicly available, real-world video datasets with annotated visual attention exist in a driving context. DR(eye)VE~\cite{palazzi2018predicting} and BDD-A~\cite{xia2018predicting} are the most well-known large-scale annotated datasets in naturalistic and in-lab driving settings, respectively. Importantly, the recently-released annotated driving attention dataset with in-lab settings, DADA-2000~\cite{fang2019dada}, is the only available dataset capturing scenes of collisions. This is because it is nearly impossible to collect enough driver attention data for collision or near-collision events. EyeCar further contributes to this area by having a more diverse array of driving events, beyond looking forward and lane-keeping. Unlike DADA-2000, EyeCar captures collisions from a collision point-of-view~(POV) perspective~(egocentric) where the ego-vehicle is involved in the accident. Table~\ref{tbl:data-recap} compares EyeCar with similar datasets (more details in Sec.~\ref{sec:eyecar}).
\input{tables/data-recap}

%% file: tables/data-recap.tex
\begin{table}[t!]
\centering
\resizebox{\columnwidth}{!}{%
\begin{tabular}{|l||c|c|c|c|c|c|c|}
\hline
 Dataset & collision & collision-POV & speed &GPS &\#vehicles &\#frames & \#gaze\\
 \hline \hline
 DR(eye)Ve &\xmark & \xmark & \cmark& \cmark& 1.0 & 555k  & 8\\
  \hline
 BDD-A &\xmark & \xmark & \cmark& \xmark& 4.4 & 318k  & 45\\
 \hline
 DADA-2000 &\cmark & \xmark & \xmark& \xmark& 2.1 &  658k  & 20\\
  \hline
 \textbf{EyeCar} &\cmark & \cmark & \cmark& \cmark& 4.6 &  315k  & 20\\
\hline
\end{tabular}
}

\vspace{-0.1in}
\caption{Compared to prior datasets, EyeCar is the only dataset that captured collisions from a point-of-view (POV) perspective and the host vehicle is involved in the collision. Previous datasets either did not capture attention from a collision point of view or had a less crowded scene. Note that \#vehicles refer to the average number of vehicles per frame.\vspace{-0.25in}}
\label{tbl:data-recap}
\end{table}

%% file: sections/method.tex
We propose MEDIRL for predicting drivers' visual attention in accident prone situations from driving videos paired with their eye fixations. MEDIRL learns a visual attention policy from demonstrated attention behavior. We formulate the problem as the learning of a policy function that models the eye fixations as a sequence of decisions made by an agent. 
Each fixation pattern is predicted given the present agent state and the current observed world configuration~(i.e., a scene context). 


\subsection{Overview and Preliminaries}
In this section, we introduce our notation and describe the features used in our proposed model.

\begin{figure}[t]
    \centering
     \includegraphics[width=\columnwidth]{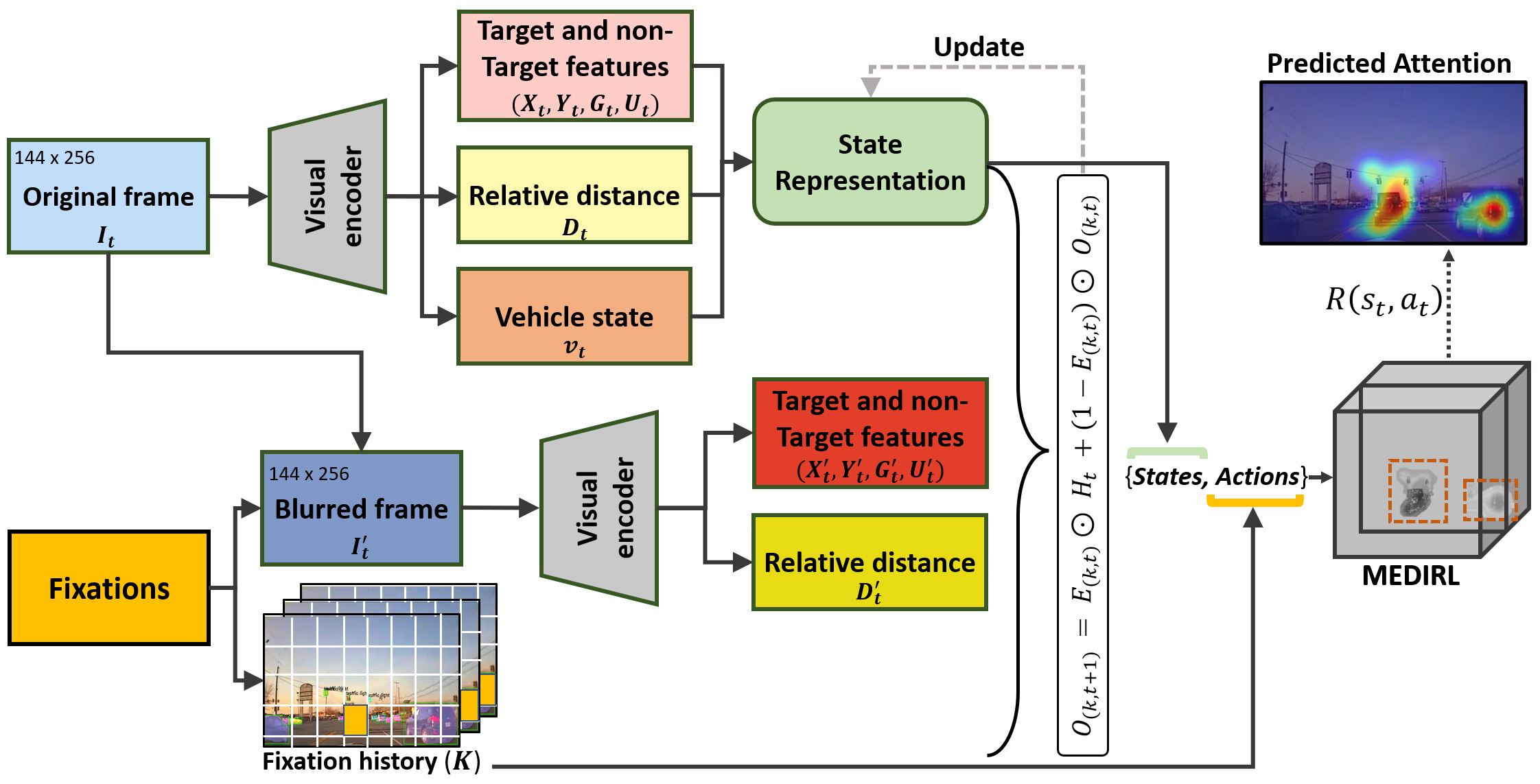}
     \vspace{-0.25in}
    \caption{Overview of our state-representation. To simulate human fovea, the agent receives high-resolution information surrounding the attended location, and low-resolution information outside of this simulated fovea. At each fixation point, a new state is generated by applying Eq.~\ref{eq:update}.
    }
    \label{fig:state-representation}
    \vspace{-0.2in}
\end{figure}

\noindent{\textbf{Visual Information.}}~During attention allocation in a dynamic and complex scene, relevant anchor objects--those with a spatial relationship to the target object--can guide attention to a faster reaction time, less scene coverage, and less time between fixating on the anchor and the target object~\cite{vo2019reading,helbing2020search,beitner2021get}. Therefore, we need to encode each frame of a given video to extract target and non-target features which an agent needs in order to effectively select the next fixation locations. Next, we describe in detail how this encoding is done~(see Figure.~\ref{fig:state-representation}). An overview of the visual encoder function is also outlined in Algorithm~\ref{alg:feature-algorithm}.

Given a family of driving video frame input, $I = \big\{I_t\big\}_{t=1}^T$, where $T$ is the number of frames. We extract visual information in a discriminative way while keeping the relevant spatial information. Each frame has several fixation locations that are processed sequentially. At each step, we extracts visual features from the current input frame. To well represent a given video frame input to an agent, we consider both pixel- and instance-level representation~(see Figure~\ref{fig:problem-setup}). The pixel-level representation determines the overall scene category by putting emphasis on understanding its global properties. The instance-level representation identifies the individual constituent parts of a whole scene as well as their interrelations on a more local instance-level. 

For pixel-level representations, we extract features~$X_t$ from a given video frame~(e.g., cars, trees). The feature extractor output is a tensor $X_t \in \mathcal{R}^{h\times w\times d}$, where h, w, and d are the height, width, and channel, respectively. At the instance-level, we represent the bounding box or instance-mask to reason explicitly over instances~(e.g., lead-vehicle) rather than reasoning over all objects representation. We utilize a position-sensitive ROI average pooling layer~\cite{yang2019video} to extract region features~$Y_t$ for each box.

To extract features relevant to a driving task, we also consider the road lanes along with the lead vehicle features in our visual representation. The road lanes~($G_t$) are critical for the task-related visual attention of drivers since they are an important indicator of the type of maneuver~\cite{do2017human}. To amplify the predicted attention for pixels of the target objects, we detect the lead vehicle~($M_t$) which is important in rear-end collisions~\cite{lyu2020towards}. The lead vehicle is a critical anchor object that can direct the driver attention to the target object, i.e. brake lights. We discretize each frame into an $n\times m$ grid where each patch matches the smallest~(furthest) size of the lead vehicle bounding box~(see Figure.~\ref{fig:state-representation}). In addition, we extract pixel locations of the brake lights by first converting each frame to the HSV color space, and then using a position-sensitive ROI max-pooling layer to extract region features for the lead vehicle box~($U_t$). The boxes and their respective features are treated as a set of objects.

\noindent{\textbf{Relative Distance.}}~Drivers pay more attention to the objects which are relatively closer as opposed to those at a distance, since the chance of collision is significantly higher for the former case. Thus, relative distance between objects and the ego-vehicle is crucial for making optimal driving decisions~\cite{pal2020looking}. To amplify nearby regions of a driving scene, we use dense depth map~($D_t$) and combine it with the general visual features~($Y_t$) by using the following formula:
\vspace{-2mm}
$$Z_t = Y_t \oplus D_t = Y_t \odot \lambda*D_t + Y_t,  \lambda = 1.2 \vspace{-2mm}$$
where $\lambda$ is an amplification factor

\noindent{\textbf{Driving Tasks.}}~To discover which features of an observed environment are the most driving task related, we need to determine the types~($Q_t$) of driving task. We observed three driving tasks ending to rear-end collisions across all videos: \textit{lane-keeping}, \textit{merging-in}, and \textit{braking}. We use function~\textit{$f_{task}$} to define these driving tasks by two criteria: 1)~ego-vehicle makes lane changing decision~$c$ and 2)~the existence of a traffic signal~$I_{signal}$ in a given driving video. 
\vspace{-2.5mm}
\begin{align*}
    \text{driving task}=
    \begin{cases}
      \text{lane-keeping}, & \text{if}\ c=0,\; I_{signal} = 0 \\
      \text{merging-in}, & \text{if}\ c=1, \; I_{signal} = 0\:\text{or}\:1 \\
      \text{braking}, & \text{if}\ c=0,\; I_{signal} = 1 
    \end{cases}
\end{align*}
\vspace{-2mm}

\noindent{\textbf{Vehicle State.}}~We optionally concatenate the speed of the ego-vehicle~$v_t$, which can influence the fixation selection~\cite{yu2020bdd100k,palazzi2018predicting,pal2020looking}, with the extracted visual representation, relative distance, and driving tasks. 

\subsection{MEDIRL}
Attentive drivers predominantly attend to the task-related regions of the scene to filter out irrelevant information and ultimately make the optimal decisions. MEDIRL attempts to imitate this behavior by using the collective non-target and target features~--extracted through parsing the driving scene--~in the state representation. Subsequently, it integrates changes in the state representation with alterations in eye fixation point, to predict fixation. 
Therefore, the \textbf{state} of an agent is determined by a sequence of visual information that accumulates through fixations towards the target object~(i.e., a brake light) which we call it a foveated frame, Figure~\ref{fig:problem-setup} shows an example of a foveated frame. The \textbf{action} of an agent, the next fixation location, depends on the state at that time. The \textbf{goal} of an agent is to maximize internal \textbf{reward} by encapsulating the intended behavior of attentive drivers~(experts) through changes in fixation locations. MEDIRL employs IRL to recover this reward function~($R$) from the set of demonstrations. 

\vspace{-4mm}
\paragraph{State Representation:} MEDIRL considers the following components in the state representations: simulating the human visual system, collecting a context of spatial cues, and modeling state dynamics. See Algorithm~\ref{alg:feature-algorithm} for describing the overview of the state representation.

\textit{Human visual system~(fovea)}: Human visual system accumulates information by attending to a specific location within the field of view. Consequently, humans selectively fixate on new locations to make optimal decisions. It means high-resolution visual information is available only at a central fixated location and the visual input outside of the attend location becomes progressively more blurred with distance away from the currently fixated location~\cite{zelinsky2019benchmarking}. We simulate human fovea by capturing high-resolution information about the current fixation location and a surrounding patch with a size $12\times17$~(about 1$\degree$ visual angle), as well as low-resolution information outside of the simulated fovea~\cite{zelinsky2019benchmarking}. To effectively formulate this system, MEDIRL uses a local patch from the original frames of the video as the high-resolution foveal input and a blurred version of the frame to approximate low-resolution input~$L$ from peripheral vision~\cite{zhang2018agil}. We obtain the blurred frames by applying a Gaussian smoothing with standard deviation $\sigma = 2\times d$, which $d$ is equal to Euclidean distance between the current fixation point~$p_{k,t}$, where $k = {0,...,\mathcal{K}}$, and the size of the frame. Note that the number of fixations~$K$ varies from frame to frame. 

\textit{Spatial cues}: A driving task and the driving-relevant~(anchor) objects of the scene can potentially direct drivers' attention to the primary target object. For example, drivers consider the distance to the lead vehicle when they brake. To approximate this guided selection of fixations, MEDIRL includes visual information in the state representation. This state representation collects the non-target and target features can create a context of spatial and temporal cues that might affect the selection of drivers' fixations. 

\input{tables/algorithm}
\textit{Dynamics of state}: To model the altering of the state representation followed by each fixation, we propose a dynamic state model. To begin with, the state is a low-resolution frame corresponding to peripheral visual input. After each fixation made by a driver, we update the state by replacing the portion of the low-resolution features with the corresponding high-resolution portion obtained at each new fixation location~(see Figure.~\ref{fig:state-representation}). At a given time step $t$, feature maps~$H$ for the original frame~(high-resolution) and feature maps~$L$ for the blurred frame~(low-resolution) are combined as follows:
\vspace{-2mm}
$$O_{0,1} = L_{0,1}, O_{k+1,t} = E_{k,t}\odot H_t + (1 - E_{k,t})\odot O_{k,t}\label{eq:update},\vspace{-2mm}$$
where $\odot$ is an element-wise product. $O_{k,t}$ is a context of spatial cues after $k$ fixations. $E_{k,t}$ is the circular mask generated from the k$^{th}$ fixation~(i.e., it is a binary map with 1 at current fixation location and 0 elsewhere in a discretize frame). To jointly aggregate all the temporal information, we update the next frame by considering all context of spatial cues in the previous frame as follows:
\vspace{-3mm}
$$O_{k,t+1} = E_{k,t+1}\odot H_{t+1} + (1 - E_{k,t+1})\odot O_{\mathcal{K},t},\vspace{-3mm}$$
where $O_{\mathcal{K},t}$ is visual information after all fixations~$\mathcal{K}$ of time step $t$(previous frame).

Drivers have various visual behaviors while performing a driving tasks and many factors~(e.g. speed) may affect the chosen fixation strategy~\cite{yu2020bdd100k,palazzi2018predicting,pal2020looking}. To efficiently predict fixations for all drivers, we augment the state by aggregating it with a high-dimensional latent space that encodes the driving task~$Q_t$. We then add another fully-connected layer to encode the current speed of the ego-vehicle~$v_t$ and concatenate the state with the speed vector. With the visual information and ego-vehicle state at each time step, we fuse all into a single state. The state of the agent is then complete in the sense that it contains all bottom-up, top-down, and historical information~(more detail of these components can be found in the supplementary material). 

\vspace{-4mm}
\paragraph{Action Space:} Herein we aim to predict the next eye fixation location of a driver. 
Therefore, the policy selects one out of $n*m$ patches in a given discretize frame. The center of the selected patch in the frame is a new fixation. Finally, the changes~($\Delta_x,\Delta_y$) of the current fixation and the selected fixation define the action space~$A_t$: \{left, right, up, down, focus-inward, focus-outward, stay\}, as shown in Figure~\ref{fig:problem-setup} which has three degrees of freedom~(vertical, horizontal, diagonal). 

\input{tables/results-benchmarks}
\vspace{-4mm}
\paragraph{Reward and Policy:}To learn the reward function and policies of driver visual attention in rear-end collisions, we use a \textbf{maximum entropy} deep inverse reinforcement learning~\cite{wulfmeier2015maximum}. MEDIRL assumes the reward is a function of the state and the action, and this reward function can be jointly learned using the imitation policy. 

The main goal of IRL is to recover the unknown reward function $R$ from the set of demonstrations~$\Xi = \{\xi_1, \xi_2,...,\xi_q\}$, where $\xi_q = \{(s_1,a_1),...,(s_\tau,a_\tau)\}$. We use maximum entropy deep IRL, which models trajectories as being distributed proportional to their exponentiated return:
\vspace{-2.5mm}
$$p(\xi) = (\nicefrac{1}{Z}) exp(R(\xi)),\vspace{-1mm}$$
where $Z$ is the partition function, $Z = \int_\xi exp(R(\xi))d\xi$. To approximate the reward function, we assume it can be represented as $R = \omega^T\phi$, where $\omega$ is a weight vector and $\phi$ is a feature vector. Such representation is constrained to be linear with respect to the input features $\phi$. In order to learn a reward function with fewer constraints, we use deep learning techniques to determine $\Phi(\phi,\theta)$, a potentially higher dimensional feature space, and approximate the reward function as $R = \omega^T\Phi(\phi,\theta)(s,a)$. Note that the weight vectors of $\omega$ and the parameter vector $\theta$ are both associated with the network which is fine-tuned by jointly training the different category of driving tasks.


\vspace{-4mm}
\paragraph{Loss Function:}To learn the attention policies, MEDIRL maximizes the joint posterior distribution of fixation selection demonstrations $\Xi$, under a given reward structure and of the model parameter, $\theta$. For a single frame and a given fixation sequence $\xi$ with a length of $|\tau|$, the likelihood is:
\vspace{-2mm}
$$\mathcal{L_{\theta}} = (\nicefrac{1}{\Xi}) \sum_{\xi^{i}\in \Xi} log P(\xi^{i}, \theta),\vspace{-2.5mm}$$ 
where $P(\xi^{i}, \theta)$ is the probability of the trajectory $\xi^i$ in demonstration $\Xi$. 

The algorithm tries to select a reward function that induces an attention policy with a maximum entropy distribution over all state-action trajectories and minimum empirical Kullback-Leibler divergence~(KLD) from drivers state-action pairs. In each iteration~($q$) of maximum entropy deep IRL algorithm, we first evaluate the reward value based on the state features and the current reward network parameters~($\theta_q$). Then, we determine the current policy~($\pi_q$) based on the current approximation of reward~($R_q$), and transition matrix~$\mathcal{T}$~(i.e., the outcome state-space of a taken action). We benefit from the maximum entropy paradigm, which enables the model to handle sub-optimal and stochastic visual behavior of drivers, by operating on the distribution over possible trajectories~\cite{ziebart2008maximum, wulfmeier2015maximum}.


%% file: tables/algorithm.tex
\begin{algorithm}[htb]
  \scriptsize
\caption{MEDIRL State Representation}
\label{alg:feature-algorithm}
\begin{algorithmic}[1]
\Function{Visual Encoder}{a video frame $I$}
    \State X := \textit{HRnet}(I) \Comment{global feature}
    \State O := \textit{mask-rcnn}(I) \Comment{list of detected object}
    \State Y := \textit{ROI-average}(O, X) \Comment{extract region features}
    \State G, c := \textit{VPG-net}(I) \Comment{detect road lanes and lane changes}
    \State M, $I_{signal}$ := \textit{mask-rcnn}(Y) \Comment{detect lead-vehicle and traffic signal}
    \State U := \textit{ROI-max}(\textit{HSV-color}(I),M) \Comment{detect brake lights }
    \State D := \textit{MonoDepth2}(I) \Comment{compute relative distance} 
    \State Z := Y $\oplus$ D \Comment{amplify close objects} 
    \State Q := $f_{task}(c, I_{signal})$ \Comment{compute driving task}
    \State visual-cues = \textit{concatenate}(G, M, U, Z) \Comment{a context of spatial cues}
    \State v := ego-vehicle speed \Comment{vehicle state}
    \State \Return {\textit{visual-cues, v, Q}} \Comment{return all extracted features}
\EndFunction
\Function{blur}{frame $I$, fixation $k$}
    \State d = Euclidean($k$,size(I)) 
    \State I' = \textit{GaussianBlur($I, \sigma$)}, $\sigma = 2 \times d$ \Comment{apply a Gaussian smoothing}
    \State \Return {\textit{I'}} \Comment{return the low-resolution frame}
\EndFunction
\Procedure{State dynamics}{frame $I_t$, fixations $\mathcal{K}$}
     \For{k $\in$ K do} 
        \State \textcolor{darkgray}{\textit{\# collect context of spatial cues based on a simulated fovea movements}}
        \State $H_t$ := VisualEncoder($I_t$) 
        \State $L_{k,t}$ := VisualEncoder(blur($I_t$,k)) 
        \State \textcolor{darkgray}{\textit{\# update the state that occurs following each fixation}}
        \State $O_{0,1} = L_{0,1}$ \Comment{initialize frame corresponding to peripheral vision}
        \State \textcolor{darkgray}{\textit{\# $E_{k,t}$ is the circular mask generated from the k-th fixation}}
        \State $O_{k+1,t} = E_{k,t}\odot H_t + (1 - E_{k,t})\odot O_{k,t}$ 
    \EndFor
\EndProcedure
\end{algorithmic}
\end{algorithm}


%% file: tables/results-benchmarks.tex
\begin{table*}[htb]
\centering
\resizebox{0.80\textwidth}{!}{%
\begin{tabular}{|l|l||c|c|c|c|c|c|c|c|c|}
\hline
\parbox[t]{2mm}{\multirow{2}{*}{\rotatebox[origin=c]{90}{Data}}}&\multirow{2}{*}{\diagbox{Method}{Task}} &\multicolumn{3}{c|}{Merging-in} & \multicolumn{3}{c|}{Lane-keeping}& \multicolumn{3}{c|}{Braking}\\
\cline{3-11}
           &    &   CC$\uparrow$  &   s-AUC$\uparrow$ & KLD$\downarrow$  & CC$\uparrow$    &   s-AUC$\uparrow$ & KLD$\downarrow$  &   CC$\uparrow$  &s-AUC$\uparrow$ & KLD$\downarrow$\\\hline\hline
\parbox[t]{2mm}{\multirow{6}{*}{\rotatebox[origin=c]{90}{DR(eye)VE~\cite{palazzi2018predicting}}}}

&Multi-branch~\cite{palazzi2018predicting} & 0.48 & -  & 2.80 & 0.55 & - & 1.87 & 0.71 & - & 2.20 \\
&HWS~\cite{xia2018predicting}          & 0.51 & -  & 2.12 & 0.75 & - & 1.72 & 0.74 & - & 1.99 \\
&SAM-ResNet~\cite{cornia2018predicting}    & \textbf{0.78} & -  & 2.01 & 0.80 & - & 1.80 & 0.79 & - & 1.89 \\
&SAM-VGG~\cite{cornia2018predicting}      & \textbf{0.78} & -  & 2.05 & 0.82 & - & 1.84 & 0.80 & - & 1.81 \\
&TASED-NET~\cite{min2019tased}      & 0.68 & -  & 1.89 & 0.73 & -  & 1.71 & 0.70 & -  & 1.89 \\
\cline{2-11}
&MEDIRL (ours) & \textbf{0.78} & -  &\textbf{0.88} & \textbf{0.89} & - & \textbf{0.75} & \textbf{0.85} &- & \textbf{0.82} \\ \hline 
\parbox[t]{2mm}{\multirow{6}{*}{\rotatebox[origin=c]{90}{BDD-A~\cite{xia2018predicting}}}}

&Multi-branch~\cite{palazzi2018predicting} & 0.58 & 0.51  & 2.08 & 0.75 & 0.72 & 2.00  & 0.69 & 0.77 & 2.04 \\
&HWS~\cite{xia2018predicting}          & 0.53 & 0.59  & 1.95 & 0.67 & 0.89 & 1.52 & 0.69 & 0.81 & 1.59 \\
&SAM-ResNet~\cite{cornia2018predicting}   & 0.74 & 0.61  & 2.00 & 0.89 & 0.79 & 1.83 & 0.85 & 0.88 & 1.89 \\
&SAM-VGG~\cite{cornia2018predicting}      & 0.76 & 0.62  & 1.79 & 0.89 & 0.82 & 1.64 & 0.86 & 0.87 & 1.85 \\
&TASED-NET~\cite{min2019tased}      & 0.73 & 0.68  & 1.83 & 0.81 & 0.66  & 1.17 & 0.87 & 0.88  & 1.12 \\
\cline{2-11}
&MEDIRL (ours) & \textbf{0.82} &\textbf{ 0.79}  & \textbf{0.91} & \textbf{0.94} & \textbf{0.91} & \textbf{0.85} & \textbf{0.93} & \textbf{0.92} & \textbf{0.89 }\\ \hline 
\parbox[t]{2mm}{\multirow{6}{*}{\rotatebox[origin=c]{90}{DADA-2000~\cite{fang2019dada}}}}

&Multi-branch~\cite{palazzi2018predicting} & 0.44 & 0.53  & 3.65 & 0.69 & 0.54 & 2.85 & 0.67 & 0.64 & 2.91 \\
&HWS~\cite{xia2018predicting}        & 0.49 & 0.59  & 3.02 & 0.72 & 0.53 & 2.65 & 0.69 & 0.77 & 2.80 \\
&SAM-ResNet~\cite{cornia2018predicting}   & 0.65 & 0.61  & 2.39 & 0.78 & 0.64 & 2.32 & 0.75 & 0.81 & 2.34 \\
&SAM-VGG~\cite{cornia2018predicting}    & 0.68 & 0.60  & 2.41 & 0.76 & 0.62 & 2.24 & 0.75 & 0.80 & 2.35 \\
&TASED-NET~\cite{min2019tased}      & 0.69 & 0.66  & 1.98 & 0.78 & 0.69  & 1.87 & 0.80 & 0.81  & 1.45 \\
\cline{2-11}
&MEDIRL (ours) &\textbf{ 0.70} & \textbf{0.68}  & \textbf{1.31} & \textbf{0.89} & \textbf{0.71} & \textbf{0.92} & \textbf{0.81} & \textbf{0.88} &\textbf{ 0.99} \\ 
\hline
\end{tabular}
}
\vspace{-0.1in}
\caption{Performance comparison of driver attention prediction on benchmarks. Models trained on the BDD-A~\cite{xia2018predicting} train set and tested on Dr(eye)VE~\cite{palazzi2018predicting}, BDD-A~\cite{xia2018predicting}, and DADA-2000~\cite{fang2019dada} test sets.\vspace{-0.25in}\label{tbl:benchmarks}}
\end{table*}



%% file: sections/eyecar.tex
Attentional lapses in normal situations~(e.g., lane-following, empty road) do not cost the same as accident-prone situations~(e.g., rear-end collision) where the cost of making an error is high. Nevertheless, collecting enough eye movements from drivers in accident-prone situations is nearly impossible because they are rather uncommon. In addition, driver attention data collected in-car has two main drawbacks~\cite{xia2018predicting, xia2020periphery}: 1)~missing covert attention: eye-trackers can only record a single focus of drivers while a driver may be attending to multiple important objects, and 2)~false positive gaze: drivers can be distracted to potential disturbances~(e.g., side road advertisement) that are not relevant to the driving. Prior works~\cite{xia2018predicting,xia2020periphery} addressed these issues with in-lab data collection, collecting drivers' eye movements while performing simulated driving tasks. 

Although in-lab driver attention collection is inevitably different from in-car driver attention, BDD-A in-lab experimental protocol showed that in-lab visual attention data reliably reveal where a driver should look at and identify the potential risks. Therefore, we follow their established and standardized experimental design protocol for collecting in-lab driver attention and create the EyeCar dataset exclusively for rear-end collisions. In order to incentivize users to pay attention and play the fall-back ready role in autonomous vehicles, we further modified the experimental design by sitting them in a low-fidelity driving simulator. The simulator consisting of a Logitech G29 steering wheel, accelerator, brake pedal, and eye-tracker~(see supplementary materials for more details).

We recruited 20 participants (5 female and 15 male, ages 22-39) with at least three years of driving experience (Mean=9.7,~SD=5.8). Participants watched all 21 selected dash-cam videos~(each lasted approximately 30sec) to identify hazardous cues in rear-end collisions. The EyeCar dataset contains 3.5 hours of gaze behavior~(aggregated and raw) captured from more than 315,000 rear-end collisions video frames. Each frame comprises 4.6 vehicles on average which makes EyeCar driving scenes more complex than other visual attention datasets~(see Table~\ref{tbl:data-recap}). The extracted speed from each frame shows that 38\% of vehicles were driving high~($65\leq v$), 39\% normal~($35\leq v \leq 65$), and 23\% low~($35\geq v$). EyeCar also provides a rich set of annotations(e.g., scene tagging, object bounding, lane marking, etc.; more details in supplementary materials).

\vspace{-3mm}



\input{tables/result-on-eyecar}

%% file: tables/result-on-eyecar.tex
\vspace{.5em}
\begin{table*}[htb]
\centering
\resizebox{0.80\textwidth}{!}{%
\setlength{\tabcolsep}{5.5pt}
\begin{tabular}{|l|l||c|c|c|c|c|c|c|c|c|}
\hline
\parbox[t]{2mm}{\multirow{2}{*}{\rotatebox[origin=c]{90}{Data}}}&\multirow{2}{*}{\diagbox{Method}{Task}} &\multicolumn{3}{c|}{Merging-in} & \multicolumn{3}{c|}{Lane-keeping}& \multicolumn{3}{c|}{Braking}\\
\cline{3-11}
           &    &   CC$\uparrow$  &   s-AUC$\uparrow$ & KLD$\downarrow$  & CC$\uparrow$    &   s-AUC$\uparrow$ & KLD$\downarrow$  &   CC$\uparrow$  &s-AUC$\uparrow$ & KLD$\downarrow$\\\hline \hline
\parbox[t]{2mm}{\multirow{6}{*}{\rotatebox[origin=c]{90}{DR(eye)VE~\cite{palazzi2018predicting}}}}&Multi-branch~\cite{palazzi2018predicting} &   0.36      &   0.37     &   6.46    &   0.51 & 0.49 & 4.80 & 0.69 & 0.49 &   3.38 \\
&HWS~\cite{xia2018predicting}     &   0.38      &   0.34   &   4.38    &   0.71 & 0.51 & 4.44 & 0.72 & 0.61 &   3.30 \\
&SAM-ResNet~\cite{cornia2018predicting}   &   0.49    &   0.48   &   4.29   &   0.73 & 0.55 & 3.90  & 0.74 & 0.66 &   3.27 \\
&SAM-VGG~\cite{cornia2018predicting}   &   0.50   &   0.47     &   4.31    &  0.74 & 0.53 & 3.95 & 0.75 & 0.64 &   3.29 \\
&TASED-NET~\cite{min2019tased}      & 0.48 & 0.46  & 3.95 & 0.74 & 0.55  & 3.81 & 0.76 & 0.65  & 3.23 \\
\cline{2-11}
&MEDIRL (ours)   &  \textbf{ 0.51 } &  \textbf{ 0.51 }   &   \textbf{2.32}    &   \textbf{0.76} & \textbf{0.57} & \textbf{3.11} & \textbf{0.79} & \textbf{0.69} &   \textbf{3.07} \\ \hline 
\parbox[t]{2mm}{\multirow{6}{*}{\rotatebox[origin=c]{90}{BDD-A~\cite{xia2018predicting}}}}&Multi-branch~\cite{palazzi2018predicting} &   0.46      &   0.48 &   4.42   & 0.51 &0.61 & 3.57 & 0.61 & 0.64 &   3.08 \\
&HWS~\cite{xia2018predicting}      &   0.41      &   0.47   &   4.36    &  0.69 & 0.81 & 3.55 & 0.67 & 0.68 &   2.86 \\
&SAM-ResNet~\cite{cornia2018predicting}   &   0.55     &   0.48  &   3.85    &  0.85 & 0.72 & 3.29  & 0.79 & 0.74 &   2.46 \\
&SAM-VGG~\cite{cornia2018predicting}   &   0.53      &   \textbf{0.49} &   3.92    & 0.84  &0.70 &  3.22 & 0.77 & 0.70  &   2.49 \\
&TASED-NET~\cite{min2019tased}      & 0.55 & \textbf{0.49}  & 3.78 & 0.84 & 0.71  & 3.12 & 0.77 & 0.76  & 2.47 \\
\cline{2-11}
&MEDIRL (ours)   &   \textbf{0.58 } &   \textbf{0.49}    &   \textbf{2.81}    & \textbf{0.86}  &   \textbf{0.73}& \textbf{2.43} & \textbf{0.79} & \textbf{0.81} &   \textbf{2.30} \\ \hline
\parbox[t]{2mm}{\multirow{6}{*}{\rotatebox[origin=c]{90}{DADA-2000~\cite{fang2019dada}}}}&Multi-branch~\cite{palazzi2018predicting} &   0.21      &   0.38 &   6.46    &   0.45 & 0.44 & 4.67 & 0.54 & 0.59 &   3.12 \\
&HWS~\cite{xia2018predicting}     &   0.31      &   0.35   &   6.12    &   0.51 & 0.47 & 4.54 & 0.67 & 0.71 &   3.10 \\
&SAM-ResNet~\cite{cornia2018predicting}   &   0.33     &   0.38   &   5.28  & 0.65  &   0.56  & 4.42 & 0.77& 0.71  &   3.07 \\
&SAM-VGG~\cite{cornia2018predicting}   &   0.30      &   0.39    &   5.35    &  0.69 & 0.57 & 4.31 & 0.74& 0.69  &   3.10 \\
&TASED-NET~\cite{min2019tased}      & 0.32 & 0.38  & 4.76 & 0.68 & 0.57  & 3.99 & 0.73 & 0.74  & 3.01 \\
\cline{2-11}
&MEDIRL (ours)   &  \textbf{ 0.41}  &   \textbf{0.45}    &   \textbf{3.79}  &  \textbf{0.73} & \textbf{0.60} & \textbf{2.51} & \textbf{0.75} & \textbf{0.79} &   \textbf{2.51} \\
\hline
\end{tabular}
}
\vspace{-0.1in}
\caption{Performance comparison of driver attention prediction on EyeCar. The models trained on Dr(eye)VE~\cite{palazzi2018predicting}, BDD-A~\cite{xia2018predicting}, and DADA-2000~\cite{fang2019dada} train sets and tested on EyeCar.\vspace{-0.25in}}
\label{tbl:eyecar}
\end{table*}

%% file: sections/experiment.tex
\vspace{-1mm}
\noindent{\textbf{Training details.}}~Driver attention is often strongly biased towards the vanishing point of the road and does not regularly change in a normal driving situation~\cite{xia2018predicting,pal2020looking}. However, attentive drivers regularly shift their attention from the center of the road to capture important cues in accident-prone situations. MEDIRL aims to predict driver attention in critical situations. Thus, to learn driving task-specific fixations and to avoid a strong center bias in our model two criteria were imposed when sampling training frames: 1)~train on \textbf{important frames}, 2)~exclude driving-irrelevant objects fixation sequence. Since a driver has to attend~(fixate) to important visual cues which usually appear in critical situations, the important frames are defined as frames wherein the attention map greatly deviates from the average attention map. We use KLD to measure the difference between the attention over each video frame and the average attention map of the entire video. The average attention map of each frame is calculated by aggregating and smoothing the gaze patterns of all independent observers~\cite{deng2019drivers}. We then sample continuous sequences of six frames as the training frames where their KLD is at least 0.89. We also exclude fixation sequences with more than 40\% focus on the irrelevant objects~(e.g., trees, advertisement). 

\noindent{\textbf{Datasets.}}~We evaluate our model on three driver attention benchmark datasets: DR(eye)VE~\cite{palazzi2018predicting}, BDD-A~\cite{xia2018predicting}, DADA-2000~\cite{fang2019dada} and EyeCar. To predict driver attention related to rear-end collisions, we extract the full stopping events~(resembling near-collisions) from DR(eye)VE and BDD-A, and rear-end collision events from DADA-2000. After applying the exclusion standard, we were left with 400, 1350, and 534 events in DR(eye)VE, BDD-A, and DADA-2000, respectively. Finally, within each type of driving task, we randomly split each of them into three sets of: 70\% training, 10\% validation, and 20\% test. 

\vspace{0.25in}
\begin{figure}[t!]
    \centering
    \includegraphics[width=\linewidth]{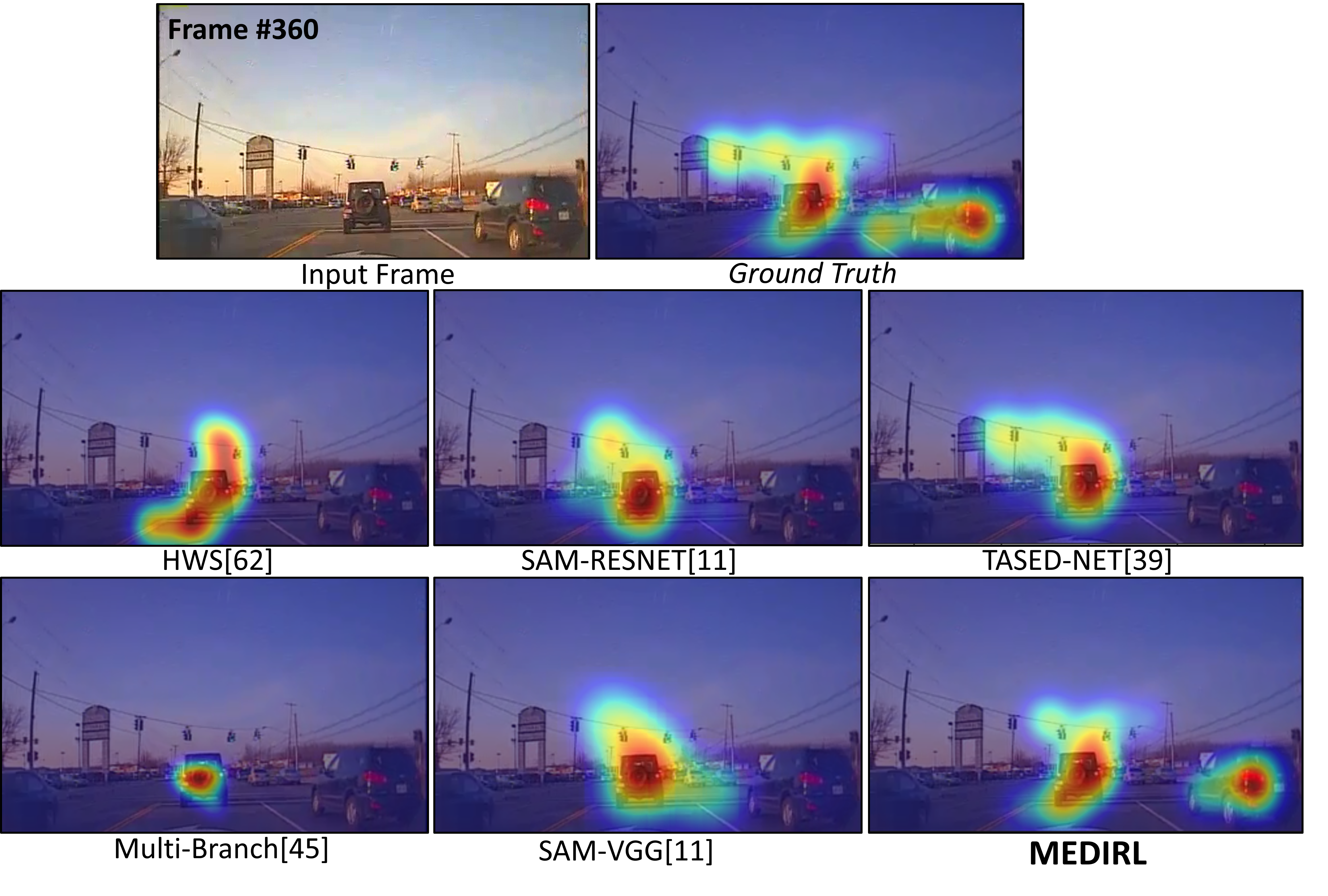}
    \vspace{-0.25in}
    \caption{Predicted driver attention in a braking task for each compared model and MEDIRL. They all trained on BDD-A. MEDIRL can learn to detect most task-related salient stimuli~(e.g., traffic light, brake light). Redder color indicates the expectation of higher reward for fixation location. More examples in supplementary materials.} \vspace{-0.25in}
    \label{fig:vaa-results}
\end{figure}

\vspace{-0.25in}
\subsection{Implementation Details}
We resize each video frame input to $144\times 256$. Then we normalize each frame by subtracting the global mean from the raw pixels and dividing by the global standard deviation. To encode visual information~(see Sec.~\ref{sec:medirl}), we use several backbones: HRNetV2~\cite{wang2020deep}--pre-trained on Mapillary Vistas street-view scene~\cite{neuhold2017mapillary}, MaskTrack-RCNN~\cite{yang2019video}--pre-trained on youtube-VIS, Monodepth2~\cite{godard2019digging}--pre-trained on KITTI 2015~\cite{geiger2012we}, and VPGNet~\cite{lee2017vpgnet}--pre-trained on VPGNet dataset. 
\input{tables/sup-ablation}

MEDIRL consists of four hidden convolutional layers with 52, 34, 20, and 20 ReLu units, respectively; followed by seven softmax units to output a final probability map. We use batch normalization after ReLu activation and set the reward discount factor to 0.98. We also set the initial learning rate to $1.5 \times 10^{-4}$, and during the first 10 epochs, we linearly increase the learning rate to $5 \times 10^{-4}$. After epoch 11, we apply a learning rate decay strategy that multiplies the learning rate by 0.25 every three epochs. For training, we use Adam optimizer~\cite{kingma2014adam}~($\beta_1 = .9, \beta_2 = .99$) and weight decay = 0. Overall, MEDIRL is trained on 36 epochs with a batch size of 20 sequences, and each sequence had six frames. The training time of MEDIRL is approximately 1.5 hours on a single NVIDIA Tesla V100 GPU and it takes about 0.08 seconds to process each frame.

\vspace{.5mm}
\noindent{\textbf{Evaluation Metrics.}}~To evaluate attention prediction, we use location-based and distribution-based saliency metrics: KLD, shuffled Area under the ROC curve~(s-AUC), and Correlation Coefficient~(CC)~\cite{bylinskii2018different}. We report s-AUC since it penalizes models with more central prediction~\cite{borji2012probabilistic,bylinskii2018different,gao2019goal}.


\vspace{-2mm}

%% file: tables/sup-ablation.tex
\begin{table*}[t!]
\centering
\resizebox{\textwidth}{!}{
\begin{tabular}{|c|l||c|c|c|c|c|c|}
\hline
\multirow{2}{*}{} & \multirow{2}{*}{\backslashbox[165mm]{Ablated versions of MEDIRL}{Dataset}} & \multicolumn{3}{c|}{EyeCar} & \multicolumn{3}{c|}{BDD-A~\cite{xia2018predicting}}\\
\cline{3-8} 
& &   CC$\uparrow$  &   KLD$\downarrow$ &$F_{\beta}\uparrow$ & CC$\uparrow$    &   KLD$\downarrow$ &  $F_{\beta}\uparrow$  \\\hline \hline
1 & global image + IRL & 0.18 & 4.21 & 0.10 & 0.22 & 4.38 & 0.12 \\\hline
2 & non target + IRL & 0.19 & 4.15 & 0.12 & 0.20 & 4.29 & 0.12 \\\hline
3 & target+non target + IRL & 0.29 & 3.51 & 0.18 & 0.36 & 3.85 & 0.25 \\\hline
4 & target+non target+distance + IRL & 0.30 & 3.62 & 0.19 & 0.38 & 3.77 & 0.27 \\\hline
5 & lead vehicle+lane + IRL & 0.30 & 3.57 & 0.23 & 0.29 & 3.51 & 0.28 \\\hline
6 & target+non target + lane+lead vehicle + IRL & 0.36 & 3.53 & 0.21 & 0.41 & 3.47 & 0.32 \\\hline
7 & target+non target+distance + lane+lead vehicle + IRL & 0.33 & 3.43 & 0.26 & 0.35 & 3.07 & 0.34 \\\hline
8 & target+non target+distance + lane+driving task + IRL & 0.51 & 3.41 & 0.31 & 0.57 & 2.18 & 0.59 \\\hline
9 & target+non target+distance + lead vehicle+driving task + IRL & 0.66 & 2.91 & 0.49 & 0.73 & 1.07 & 0.66 \\\hline
10 & target+non target+distance+lane+lead vehicle+driving task + IRL & 0.70 & 2.78 & 0.60 & 0.87 & \textbf{0.87} & 0.75  \\\hline 
11 & \textbf{MEDIRL}: target+non target+distance+lane+lead vehicle+driving task + speed + IRL & \textbf{0.74} & \textbf{2.51} & \textbf{0.61} & \textbf{0.89} & \textbf{0.88} & \textbf{0.78}\\\hline
\end{tabular}}
\vspace{-2mm}
\caption{
Quantitative evaluation of the ablated versions of MEDIRL and full MEDIRL. All models trained on BDD-A train set and tested on EyeCar and BDD-A test sets. We mask out one part by setting the map(s) to zeros at each time.\vspace{-0.25in}}
\label{tbl:ablation}
\end{table*}

%% file: sections/result.tex


Table~\ref{tbl:benchmarks} provides the \textbf{quantitative evaluation} results of MEDIRL and five baseline attention prediction models including Multi-branch~\cite{palazzi2018predicting}, HWS~\cite{xia2018predicting}, SAM-ResNet~\cite{cornia2018predicting}, SAM-VGG~\cite{cornia2018predicting}, TASED-NET~\cite{min2019tased}. 
For fair comparisons, we directly report available results released by the authors or reproduce experimental results via publicly available source codes. In this evaluation, we trained models on BDD-A and \textbf{tested on each benchmark}. The results highlight that MEDIRL surpasses almost all models under all evaluation metrics. Most significantly, our approach can effectively predict driver attention while performing various driving tasks. Although we are unable to calculate s-AUC for Dr(eye)VE as the original fixation were not reported, the results in Table~\ref{tbl:benchmarks} also indicates that the MEDIRL's superiority is not limited to a dataset. 

Further, we evaluate MEDIRL along with other attention models on EyeCar dataset, reported in Table~\ref{tbl:eyecar}. In this experiment, we \textbf{trained models on each benchmark}~(i.e., BDD-A, DR(eye)VE, DADA) and \textbf{tested on EyeCar}. 
MEDIRL performs favorably against other counterparts. However, there is a big performance gap between~Table~\ref{tbl:benchmarks} and~\ref{tbl:eyecar}, which may indicate EyeCar has different distributions. To investigate this matter, we \textbf{trained models on EyeCar} and \textbf{tested on each benchmark}. We obtained the following results; \textit{(CC : $0.89$, KLD : $0.80$)}, \textit{(CC : $0.94$, s-AUC : $0.91$, KLD : $0.85$)}, \textit{(CC : $0.85$, s-AUC : $0.77$, KLD : $0.99$)} on DR(eye)VE, BDD-A, and DADA-2000, respectively, that are average values for all types of driving tasks. These results show the effectiveness of EyeCar on representing salient regions in critical situations and also show that EyeCar attention distribution prior to accident-prone situations is more informative than benchmarks.

Figure~\ref{fig:vaa-results} shows \textbf{qualitative comparison} of MEDIRL against other models. MEDIRL can reliably capture the important visual cues in a braking task in the case of a complex frame. In contrast, nearly all other models partially capture the spatial cues and predict attention mainly towards the center of the frame, thereby ignoring the target and non-target objects~(i.e., spatial cues). Please refer to the supplementary material for more examples.


\input{sections/ablation}

%% file: sections/ablation.tex
\vspace{-0.05in}
\subsection{Ablations Studies}
To investigate how different features in our model affect its performance, we compare several ablated versions of our model against two testing sets~(i.e., EyeCar and BDD-A), using $F_{\beta}$~(${\beta}^2$ = 1~\cite{pal2020looking}), CC, and KLD. All ablated versions of our model are trained on BDD-A. 

The results show that crucial features in the model include the context of spatial cues related to target and non-target~(L3), driving-specific objects~(Line 8, 10), followed by driving task~(L9) features. MEDIRL without target~(L2) and non-target~(L5) shows a significant performance drop. From the results in Table~\ref{tbl:ablation}, we can observe that compared with the ablated versions, our full model achieves better performance, which demonstrates the necessity of each feature in our proposed model.

%% file: sections/conclusion.tex
\vspace{-1mm}           
We proposed MEDIRL, a novel inverse reinforcement learning formulation for predicting driver attention in accident-prone situations. MEDIRL effectively learns to model the fixation selection as a sequence of states and actions. MEDIRL predicts a maximally-rewarding fixation location by perceptually parsing a scene and accumulating a sequence of visual cues through fixations. To facilitate our study, we provide a new driver attention dataset comprised of rear-end collision videos with richly annotated eye information. We investigate the effectiveness of attention prediction model by experimental evaluation on three benchmarks and EyeCar. Results show that MEDIRL outperforms existing models for attention prediction and achieves state-of-the-art performance. 
\vspace{0.05in}



%% file: sections/acknowledgment.tex
\noindent{\textbf{Acknowledgements}} 
This work was supported in part by a gift from Leidos, National Science Foundation~(NSF) grant CCF-1942836, and IIS-2045773.

%% file: sections/supplementary.tex
\newcommand{\hbAppendixPrefix}{S-}
\renewcommand{\thesection}{\hbAppendixPrefix\arabic{section}}
\setcounter{section}{0}
\renewcommand{\thefigure}{\hbAppendixPrefix\arabic{figure}}
\setcounter{figure}{0}
\renewcommand{\thetable}{\hbAppendixPrefix\arabic{table}} 
\setcounter{table}{0}

\section{Supplementary Material}
We propose a novel inverse reinforcement learning formulation using Maximum Entropy Deep Inverse Reinforcement Learning (MEDIRL) for predicting the visual attention of drivers in accident-prone situations. In addition, we introduce EyeCar, a new driver attention dataset in accident-prone situations. 

In this document, we provide more details to the main paper and show extra results on ablation studies.
We provide further details about the EyeCar dataset in Section~\ref{sup:EyeCar}, and more details on the architecture and implementation of MEDIRL in Section~\ref{sup:implementation}. We also provide additional results from experiments and ablation studies~(Section~\ref{sup:ablation}). You can find the code and dataset in our Github repository\footnote{\url{https://github.com/soniabaee/MEDIRL-EyeCar}}.

\section{EyeCar}\label{sup:EyeCar}
EyeCar is a new driver attention allocation in accident-prone situations. We follow BDD-A and DADA established and standardized experimental design protocol for collecting in-lab driver attention and create the EyeCar dataset exclusively for various driving tasks which end in rear-end collisions. EyeCar covers more realistic and diverse driving scenarios in accident-prone situations. Unlike DADA-2000, EyeCar captures collisions from a collision point-of-view~(POV) perspective~(egocentric) where the ego-vehicle is involved in the accident.
\paragraph{Participants:} 
We recruited 20 participants, 5 of them were women, and the rest were men with at least three years of driving experience. You can find more details of our participants in Table~\ref{sup-tbl:participants}. Participants watched all the selected dash-cam videos to identify hazardous cues in rear-end collisions. \vspace{-0.1in}
\input{tables/sup-participants}

\paragraph{Driving videos:} We selected 21 front-view videos from the naturalistic driving dataset~\cite{dingus2015naturalistic} that included rear-end collisions with high traffic density. The videos were captured in various driving conditions. These conditions contain: traffic conditions~(e.g., crowded and not crowded), weather conditions~(e.g., rainy and sunny), landscapes~(e.g., town and highway), and times of the day~(e.g., morning, evening, night). It also contains typical driving tasks~(e.g., lane-keeping, merging-in, and braking) ending to rear-end collisions. Each rear-end collision video lasted for 30 seconds, had a resolution of 1280$\times$720 pixels, and had a frame rate of 30 frames per second. All the conditions were counterbalanced among all the participants.

\input{tables/sup-dataset}

\paragraph{Apparatus:} We conducted this study in an experiment booth with controlled lighting. The experiment was designed to maximize the accuracy of the eye tracker to be used as the ground truth for the evaluation of the estimated driver attention allocation. The driving scenes were displayed on a $20$-inch monitor with a pixel resolution of 2560 by 1440. Participants were seated approximately 60~cm away from the screen. The head was stabilized with a chin and forehead rest. A Logitech G29 steering wheel is placed in front of the participants who were asked to view the videos by assuming that they were driving a car. To control the lighting and minimize possible shadows, a Litepanels LED-daylight was used.

\input{tables/sup-videos}
\begin{figure*}[t!]
    \centering
    \includegraphics[width=0.85\textwidth]{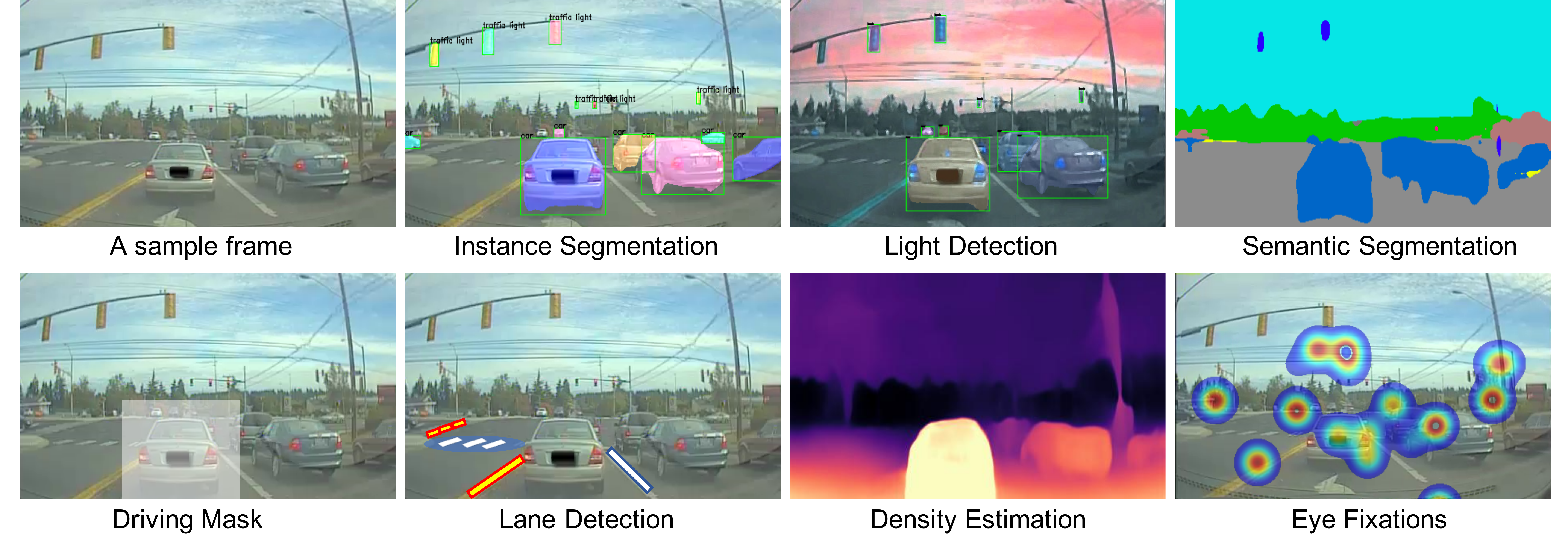}
    \vspace{-0.2in}
    \caption{Overview of our dataset. The dataset also comes with a rich set of annotations: object bounding, lane marking, full-frame semantic, and instance segmentation.\vspace{-0.2in}}
    \label{sup-fig:semantic}
\end{figure*}
Eye movements were recorded using the screen-mounted Tobii X3-120 system with a sampling rate of 120~Hz. The eye-tracker was mounted under the screen of the monitor placed in front of the participants. Due to the sensitivity of the eye tracker, the vertical placement of the screen was adjusted such that the center of the screen was at eye-level for each participant. The system had to be calibrated for each participant using the Tobii Pro Studio animating nine calibration points. Calibration accuracy was then recorded to be within 0.6 degrees of visual angle for both axes of all participants. 
\paragraph{Procedure:} Individuals are eligible to participate in this study if they have normal or corrected to normal vision and have at least three years of driving experience. After enrolling in the program, individuals are asked to fill out initial questions consisting of their age, driving experience, gender, whether they have experience with the semi-autonomous vehicle or not, and if they have been involved in any car accidents or not~(see Table~\ref{sup-tbl:participants}). 

The study had two sessions, and each lasts for 10$\pm$2 minutes. To decrease the chance of drivers' fatigue and disengagement, participants watched the first 10 videos in the first session, and then after 5 minutes gap, they watched the other 11 videos in the second session~(the whole study takes less than half an hour). The experiment received ethical approval from the University's Institutional Review Board. 

During the data collection, we asked participants to `task-view' the collision videos and were free to fix their eyes on their areas of interest. To incentivize participants to pay attention and play the fall-back ready role in autonomous vehicles, we further modified the experimental design by sitting them in a low-fidelity driving simulator consists of a Logitech G29 steering wheel, accelerator, brake pedal, and eye-tracker. 

\subsection{Data Preprocessing:}
\begin{figure}
    \centering
    \includegraphics[width=0.75\linewidth]{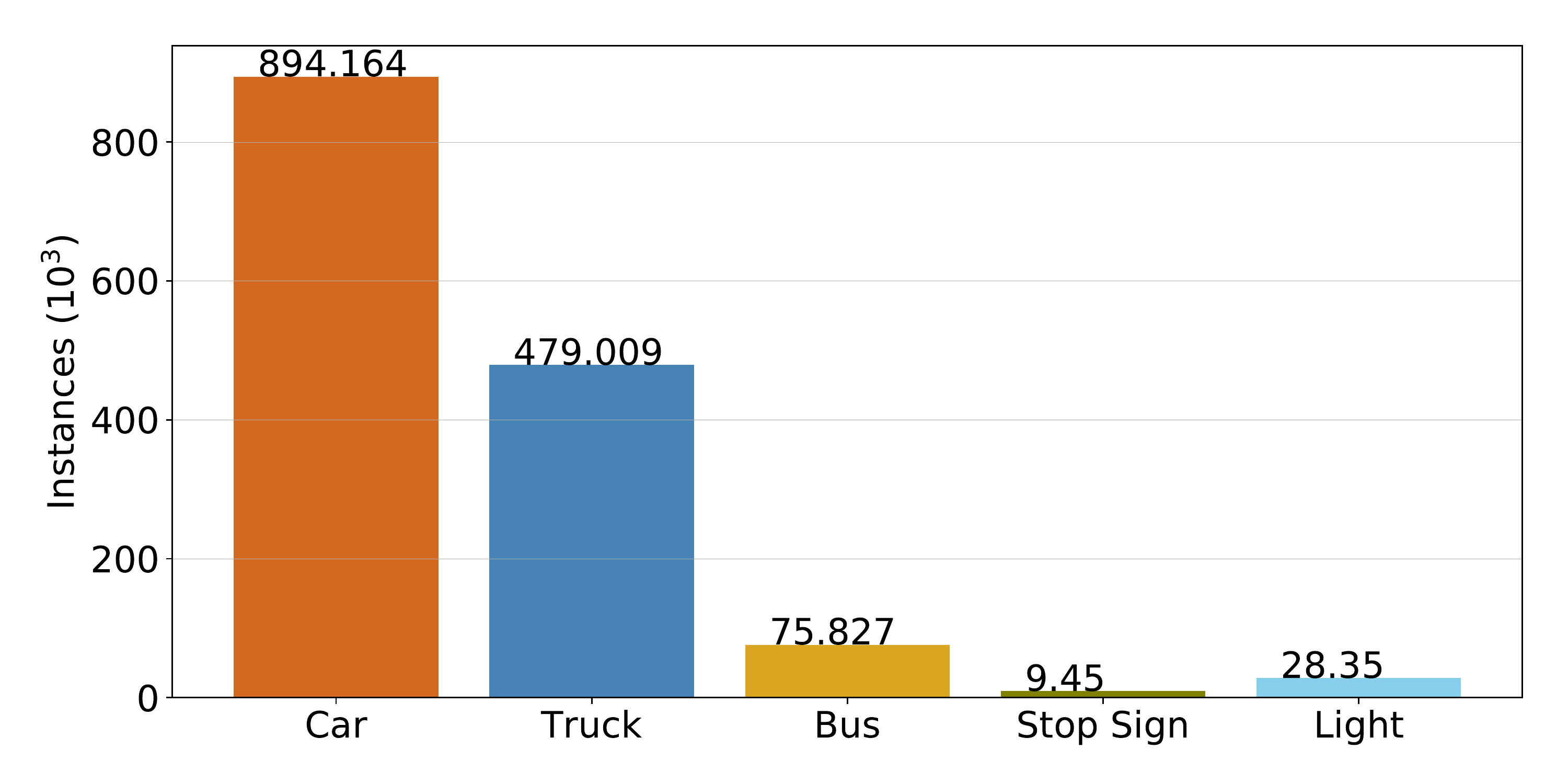}
    \caption{The distribution of typical instances categories involved in an accident over all frames of the EyeCar videos.\vspace{-0.2in}}
    \label{sup-fig:instances}
\end{figure}

\paragraph{Driving videos:}
EyeCar comes with a rich set of annotations: object bounding, lane marking, full-frame semantic, and instance segmentation~(see Figure~\ref{sup-fig:semantic}). You also can see the number of typical instances in each category involved in an accident over all frames of videos in Figure~\ref{sup-fig:instances}.

\noindent{\textbf{Object detection:}} Understanding the scene is important not only for autonomous driving but the general visual recognition. One of the main elements for a scene is the objects of the scene, therefore locating object is a fundamental task in scene understanding. We provide bounding box and instance mask annotations for each of the frames in EyeCar. The sample of the instances and masks are presented in Figure~\ref{sup-fig:semantic}. In addition, we provide the instance statistics of our object categories in Figure~\ref{sup-fig:instances}.

\noindent{\textbf{Light Detection:}} Any rear-end collision includes salient stimuli such as brake lights. To detect this type of stimuli, we convert each frame to HSV color space. First, we calculate the average brightness level of each vehicle and traffic light masks. Then, we calculate the brightness anomaly of the selected masks by subtracting their average brightness value from their actual brightness level at each frame. We can determine the pixels corresponding to these anomalies as well as their time of occurrence. Therefore, the location of the target objects and their temporal occurrence interval are annotated in EyeCar.

\noindent{\textbf{Depth Estimation:}} Recognizing the relative distance to the other traffic participants~(e.g., the lead vehicle) is crucial for making optimal driving decisions. Therefore, we use a supervised monocular depth estimation model to amplify nearby regions~(e.g., distance to a target object) of a driving scene. 

\noindent{\textbf{Lane Marking and Lane Changes:}} The lane marking detection is critical for a task-related visual attention allocation of drivers, as an indicator of the type of maneuver. We recognize the left and right lanes of the ego-vehicle by delineating their boundaries. Our lane markings (Figure~\ref{sup-fig:semantic}) are labeled with five main categories: road curb, double white, double yellow, single white, single yellow. The other categories are ignored during evaluation. 

\begin{figure*}[t!]
    \centering
    \includegraphics[width=\linewidth]{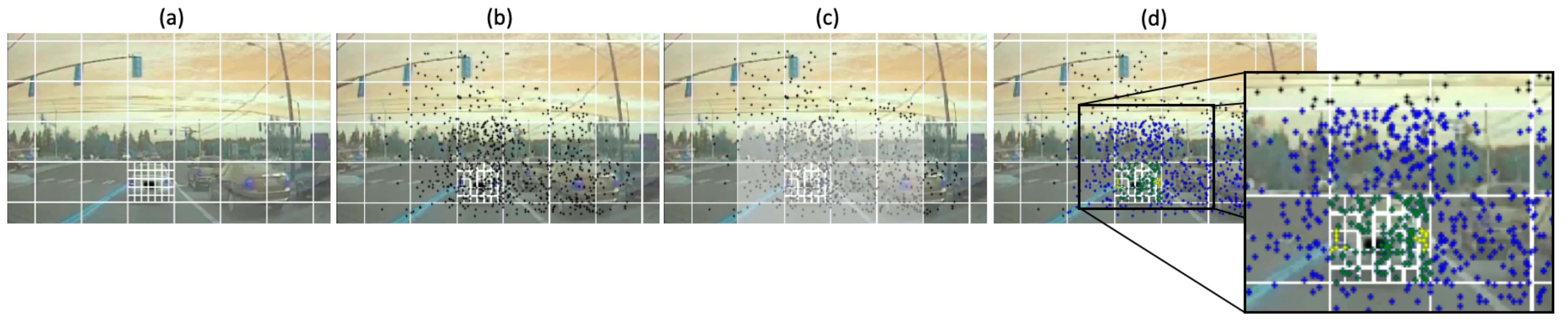}
    \caption{Illustration of a discretize frame along with gaze location points of all drivers in the EyeCar dataset. Drivers allocate their attention to the driving task-related salient regions of the driving scene. The points show the gaze location of drivers. The black points are out of the task-related regions. The blue points are in the driving mask~(the gray area in the frame). The green points are in the lead vehicle bounding box, and the yellow points are in the area of the target object~(i.e., braking lights).\vspace{-0.2in}}
    \label{sup-fig:grid}
\end{figure*}

\paragraph{Eye information:} We employ iMotion to extract the eyes' features such as; pupil size, gaze location, fixation duration, the sequence of fixation, and the start and end time of the fixation points. Moreover, the visual responses and time delay between the onset of the tasks' stimuli~(e.g., brake lights) to perceive it by the participants were captured. The abnormal or missing values of these features can lead us to the wrong conclusion, therefore pre-processing of the raw data is necessary for identifying such values and replacing them with linearly interpolated values, outlier treatment, statistical analysis, and data quality (e.g., calibration, exclusion of trials and participants due to poor recording, track loss).

\begin{figure}[t!]
    \centering
    \includegraphics[width=\linewidth]{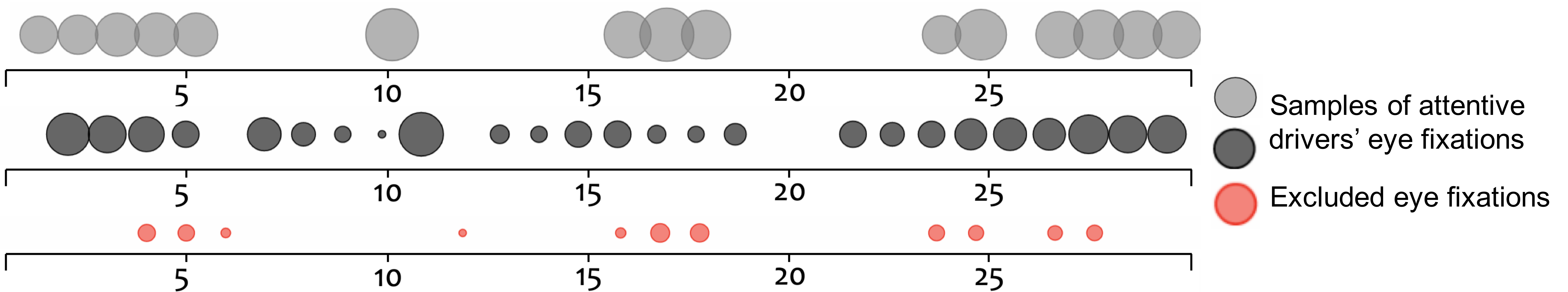}
    \caption{A sample of attentive drivers' average eye fixations sequence for a given front-view video as well as the excluded sequence. Note that the sizes of the circles are corresponding to the duration of the average eye fixation over the last 30 frames before a collision.\vspace{-0.2in}}
    \label{sup-fig:eye-fixation}
\end{figure}

To clean the data, we first extract the missing values of the eyes' features. We employ linear interpolation if the percentage of the missing values is less than 20\%. Then, we calculated the abnormal values of features to detect the outliers. We calculated the mean~($\mu_{\text{feature}}$) and the standard deviation~($\sigma_{\text{features}}$) of each feature (zero values are excluded from our calculation) for each participant. Then, we set the low and high thresholds as follows:
\begin{align*}
    \text{low threshold} = \mu_{\text{feature}} - 3 \times \sigma_{\text{feature}} \\
    \text{high threshold} = \mu_{\text{feature}} + 3 \times \sigma_{\text{feature}}
\end{align*}
Abnormal values are those that values are less than the low threshold and more than a high threshold. In addition to the exclusion criteria described in the main text, we also excluded the sequences with more than 40\% abnormal values for eye fixations~(see Figure~\ref{sup-fig:eye-fixation} for a sample of eye fixation sequence). In this way, we decreased the chance of drivers' fatigue and disengagement. We have about 0.005\% of sequences with these conditions.

\paragraph{EyeCar dataset:} After implemented all exclusion criteria, we selected 416 variable-length eye fixation sequences. EyeCar includes more than 315,000, rear-end collisions video frames. In addition, each video frame comprises 4.6 vehicles on average, making EyeCar driving scenes more complex than other visual attention datasets. GPS recordings in our dataset show the human driver action given the visual input and the driving trajectories.  The proportion of high~($65\leq v$), normal~($35\leq v \leq 65$), and low~($35\geq v$)-speed categories are 38\%, 39\%, and 23\%, respectively see Table~\ref{sup-tbl:videos}. 

A total of 1,823,159 fixations were extracted from the eye position data, over the 20 subjects. The EyeCar dataset contains 3.5 hours of gaze behavior from the 20 participants. The fixation maps highlight the direction of human drivers' gaze to a salient object when making driving decisions in rear-end collisions. We also provide a raw fixation map of multiple observers as well as an average fixation map of them. We aggregate and smooth the gaze patterns of these independent observers to make an attention map for each frame of the video~\cite{deng2019drivers} and simulate the peripheral vision of human~\cite{pal2020looking}.

\section{Implementation}\label{sup:implementation}
\paragraph{\textbf{Depth Estimation:}} The predicted dense depth map~$D_t$ at each time step~$t$ is combined with the visual feature $F_t$ by the following formula: 
$$F_t \oplus D_t = F_t \odot \lambda*D_t + F_t,$$
where $\lambda = 1.2$. This value of $lambda$ parameter helped us to focus on the lead vehicle more than other surrounding vehicles during rear-end collisions. Note that the above equation is equivalent to the main paper's equation which is written in a recurrent form.

\paragraph{\textbf{State Representation:}} In our proposed state representation, we try to formulate the visual system mechanism by considering the high-resolution visual information at the eye fixation location (a selected patch in a grid space) and low-resolution visual information outside of the eye-fixation location. 
\begin{figure}[t!]
    \centering
    \includegraphics[width=\linewidth]{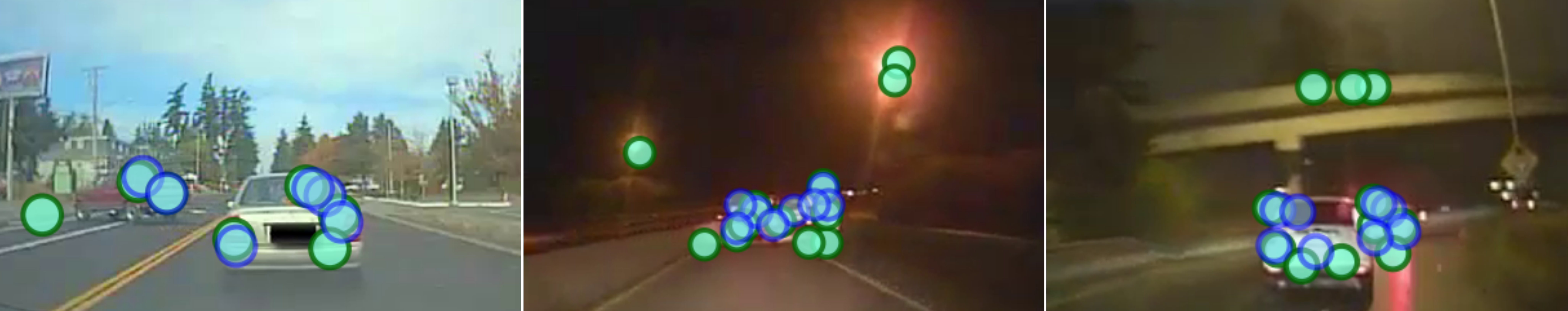}
    \caption{Examples of MEDIRL generated visual attention allocations on the EyeCar dataset. An attentive driver eye fixation sequences are colored in green, and the model generated is in blue. You can see that MEDIRL mainly focused on driving tasks related to rear-end collisions.\vspace{-0.1in}}
    \label{sup-fig:sequence}
\end{figure}

To model the altering of the state representation followed by each fixation, we propose a dynamic state model. To begin with, the state is a low-resolution frame corresponding to peripheral visual input. After each fixation made by a driver, we update the state by replacing the portion of the low-resolution features with the corresponding high-resolution portion obtained at each new fixation location. At a given time step $t$, feature maps~$H$ for the original frame~(high-resolution) and feature maps~$L$ for the blurred frame~(low-resolution) are combined as follows:
$$O_{0,1} = L_{0,1}, O_{k+1,t} = E_{k,t}\odot H_t + (1 - E_{k,t})\odot O_{k,t}\label{eq:update},$$
where $\odot$ is an element-wise product. $O_{k,t}$ is a context of spatial cues after $k$ fixations. $E_{k,t}$ is a binary map with 1 at current fixation location and 0 elsewhere in a discretize frame. The size of each patch is equal to the smallest size(furthest) of the lead vehicle in the scene $12\times17$~(about 1$\degree$ visual angle). To jointly aggregate all the temporal information, we update the next frame by considering all context of spatial cues in the previous frame as follows:
$$O_{k,t+1} = E_{k,t+1}\odot H_{t+1} + (1 - E_{k,t+1})\odot O_{\mathcal{K},t},$$
where $O_{\mathcal{K},t}$ is visual information after all fixations~$\mathcal{K}$ of time step $t$(previous frame).
\begin{figure}[h!]
    \centering
    \includegraphics[width=\linewidth]{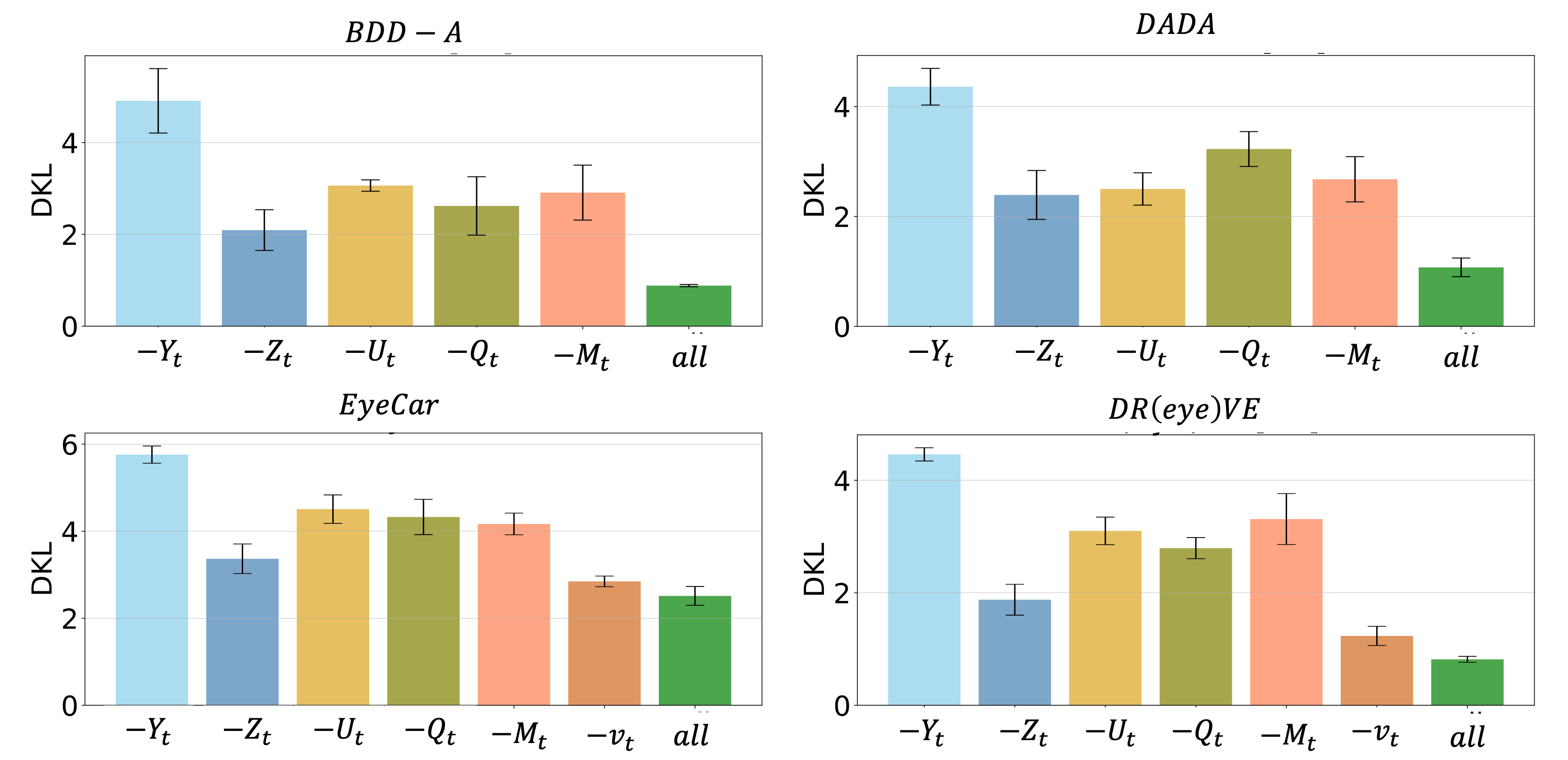}
    \caption{Ablation study on the proposed state representation. We remove one part by masking out or simply removing from the state representation at each time.\vspace{-0.1in}}
    \label{fig:ablation-state}
\end{figure}{} 
\begin{figure*}[h!]
    \centering
    \includegraphics[width=\linewidth]{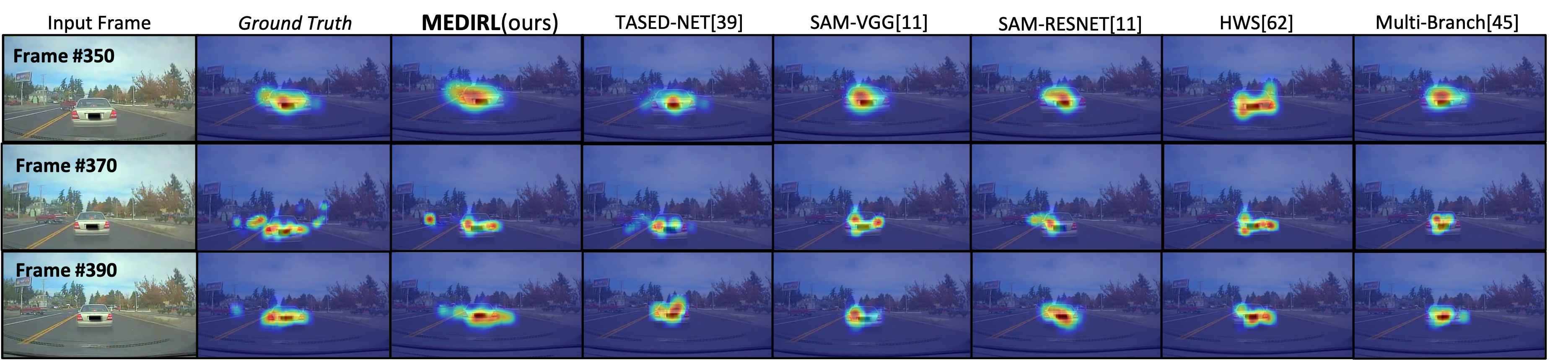}
    \caption{Predicted driver attention in a braking task for each compared model and MEDIRL (video \#17). They all trained on BDD-A. MEDIRL can learn to detect most task-related salient stimuli~(e.g., traffic light, brake light). The redder color indicates the expectation of a higher reward for fixation location.\vspace{-0.1in}}
    \label{fig:ablation-comparison}
\end{figure*}{} 
\paragraph{\textbf{Action Space:}} We aim to predict the next eye fixation of drivers. It means, we need to predict the pixel location where the driver is looking in the driving scene during accident-prone situations. We discretize each frame into a $n\times m$ grid where each patch matches the smallest size~(furthest) of the lead vehicle bounding box~(see Figure~\ref{sup-fig:grid}. The maximum approximation error due to this discretization procedure is 1.27 degrees, visual angel. Action $a_{k,t}$ represents where the focus of attention can move at fixation~$k$ of time step $t$. The policy selects one out of $n*m$ patches in a given discretize frame. The center of the selected patch in the frame is a new fixation. Finally, the changes~($\Delta_x,\Delta_y$) of the current fixation and the selected fixation define the action space~$A_t$: \{left, right, up, down, focus-inward, focus-outward, stay\}, as shown in Figure~\ref{fig:problem-setup} which has three degrees of freedom~(vertical, horizontal, diagonal). We also excluded the patches that have no visits or less than five visits for computational efficiency. It should be noted that we did not pre-defined the radius of the direction for the agent. Therefore, the agent has the freedom to pick any patch among the created ones. 

\paragraph{\textbf{Driving task and ego-vehicle speed:}} We embed the task in our model by one-hot encoding maps which spatially repeat the one-hot vector. Therefore, we concatenate the task embedding with other features in our proposed state representation to have a task-dependent bias term for every convolutional layer. We then add another fully-connected layer to encode the current speed of the ego-vehicle and concatenate the state with the speed vector.

\paragraph{\textbf{Visual attention allocations:}} The eye fixation location is generated from the probability map that MEDIRL has produced. We also applied Inhibition-of-Return to decrease the likelihood that a previously inspected~(possibly salient) region in the scene will be re-inspected, thereby encouraging visual attention toward the next salient region in a driving scene. Therefore, MEDIRL generates a new spatial probability map at every step. 

Figure~\ref{sup-fig:sequence} demonstrates the generated fixation location in a single frame of different driving videos by MEDIRL. MEDIRL mainly focused on driving tasks related to rear-end collisions. 

\paragraph{\textbf{Maximum Entropy:}} To learn the policies, we maximize the joint posterior distribution of visual attention allocation demonstrations~$\Xi = \{\xi_1, \xi_2,...,\xi_q\}$, under a given reward structure and of the model parameter, $\theta$, across $I$. For a single frame and given visual attention allocation sequence $\xi_q = \{(s_1,a_1),...,(s_\tau,a_\tau)\}$ with a length of $|\tau|$, the likelihood is: 
$$\mathcal{L_{\theta}} = (\nicefrac{1}{\Xi}) \sum_{\xi^{i}\in \Xi} log P(\xi^{i}, \theta),$$ 
, where $P(\xi^{i}, \theta)$ is the probability of the trajectory $\xi^i$ in demonstration $\Xi$. 
In each iteration~$j$ of maximum entropy deep inverse reinforcement learning algorithm, we first evaluate the reward value based on the state features and the current reward network parameters~$\theta_j$. Then, we determine the current policy, $\pi_j$, based on the current approximation of reward, $R_j$ and transition matrix~(i.e., the outcome state-space of a taken action), $\mathcal{T}$. Therefore, we can benefit from the maximum entropy paradigm, which enables the model to handle sub-optimal behavior as well as stochastic behavior of experts, by operating on the distribution over possible trajectories~\cite{ziebart2008maximum, wulfmeier2015maximum}.

Principle of Maximum Entropy~\cite{jaynes1957information} demonstrates that the best distribution overcurrent information is one with the largest entropy. Maximum Entropy also prevents issues with label bias which means portions of state space with many branches will each be biased to be less likely, and while areas with fewer branches will have higher probabilities (locally greedy). Maximum Entropy gives all paths equal probability due to equal reward and uses a probabilistic approach that maximizes the entropy of the actions, allowing a principled way to handle noise, and it prevents label bias. It also provides an efficient algorithm to compute empirical feature count, leading to a state-of-the-art performance at the time. This process maximized total reward, even over the short period of time~(0.6$\pm$ 0.2 seconds) that our attentive drivers detect the target objects~(brake light) in rear-end collisions.

\section{More Evaluations}
\subsection{Training and Testing on EyeCar}
We train and test MEDIRL on EyeCar. To be able to do it, we used leave-one-out cross-validation (one video as test) and obtained the following results: (CC: 0.85, s-AUC: 0.8, KLD: 0.92), (CC: 0.83, s-AUC: 0.74, KLD: 1.58), (CC: 0.79, s-AUC: 0.77, KLD: 1.29), on lane-keeping, merging-in, and braking driving tasks, respectively.

\subsection{Training on EyeCar and Testing on Benchmarks}
We report the results of training on EyeCar and testing on each benchmark for each driving task.

\begin{table}[H]
\centering
\resizebox{\columnwidth}{!}{%
\begin{tabular}{|l||c|c|c|c|c|c|c|c|c|}
\hline
\multirow{2}{*}{\diagbox{Dataset}{Task}} & \multicolumn{3}{c|}{Merging-in} & \multicolumn{3}{c|}{Lane-keeping} & \multicolumn{3}{c|}{Braking}\\\cline{2-10}
 & CC$\uparrow$  &   s-AUC$\uparrow$ & KLD$\downarrow$  &  CC$\uparrow$    &   s-AUC$\uparrow$ & KLD$\downarrow$  & CC$\uparrow$  &s-AUC$\uparrow$ & KLD$\downarrow$\\
\hline \hline
DR(eye)VE &  0.88 & - & 0.89 &  0.91 & -  & 0.70 & 0.88 &  - & 0.81\\\hline
BDD-A & 0.92 & 0.89 & 0.87 & 0.94 & 0.94 & 0.82  & 0.96 & 0.90 & 0.86 \\\hline
DADA-2000 & 0.77 & 0.71 & 1.06 & 0.93 & 0.72 & 0.92 & 0.85 & 0.88 & 0.99 \\ 
\hline
\end{tabular}}
\vspace{-0.1in}
\caption{The results of training on EyeCar and testing on each benchmark for each driving task.}
\end{table}

\section{Qualitative Comparison}
We provide a qualitative comparison of MEDIRL against other models in Figure~\ref{fig:ablation-comparison}. It shows that MEDIRL can reliably manage to capture the important visual cues in a braking task in the case of a complex frame. In contrast, nearly all other models partially capture the spatial cues and predict attention mainly towards the center of the frame, thereby ignoring the target and non-target objects~(i.e., spatial cues). 

\subsection{Challenging Environment}
We also evaluate MEDIRL performance under extreme weather conditions such as foggy weather. The BDD-A dataset includes severe weather (e.g., foggy and snowy). We report the results of MEDIRL trained and tested on BDD-A in Table 2 of the paper, showing MEDIRL surpasses almost all the models. We further compared MEDIRL with TASED-NET exclusively on the \textbf{foggy} videos extracted from BDD-A.

\begin{table}[H]
\tiny
\small\addtolength{\tabcolsep}{-.2pt}
\centering
\resizebox{0.9\columnwidth}{!}{%
\begin{tabular}{|l||c|c|c|c|}
\hline
  Weather & Methods & CC$\uparrow$ & s-AUC$\uparrow$ & KLD$\downarrow$\\\hline \hline
 \multirow{2}{*}{Foggy} & TASED-NET  & 0.64 & 0.53 & 2.12 \\\cline{2-5}
                        & MEDIRL  & 0.72 & 0.62 & 1.45\\
  \hline
\end{tabular}}
\vspace{-0.1in}
\caption{Evaluating MEDIRL in a challenging driving environment (i.e., foggy).}
\end{table}

Despite the foggy weather conditions, the results highlight that MEDIRL still performs better under all evaluation metrics. The results will be added to supplementary material. Generally, MEDIRL is more sensitive to false-negative prediction, leading to significant improvement in KLD.

\section{Ablation Studies}\label{sup:ablation}
Figure~\ref{fig:ablation-state} shows the ablation study of the full state representation on different test datasets. We can see that the most important feature categories were semantic/instance~($Y_t$), followed by target object~($U_t$), and types of driving tasks~($Q_t$). The depth map features~($Z_t$) are also beneficial for the model performance whereas ego-vehicle speed~($v_t$) weakly impacted model performances. The results confirm the incorporation of low and mid-level visual cues, and driving-specific visual features.

We study the benefits of each component of MEDIRL by running ablation experiments~(see Table~\ref{sup:ablation}) with the trained model on the BDD-A dataset and tested on EyeCar and BDD-A test dataset. We employed our general scene features and driving-related features to have a rich state representation for our proposed MEDIRL model. To understand the contribution of each component, we removed the maps of each group one at a time and compared the corresponding performance of the model. MEDIRL is not restricted to these backbones and could potentially incorporate new and more robust networks as submodules.

\input{tables/ablation}


%% file: tables/sup-participants.tex
\begin{table}[htbp]
\centering 
\resizebox{\linewidth}{!}{%
\begin{tabular}{|c c c c c|}
  \hline 
  \textbf{Gender} & \textbf{Age} & \textbf{Driving Experience} & \textbf{Semi-autonomous vehicle} & \textbf{Accident}\\
  \hline
  \hline
  5 female, 15 male  & 22-39 & 9.71($\pm$ 5.8) & 25\% & 1\% \\
 \hline 
\end{tabular}
}
\caption{Detailed information about individuals who participate in the study.}
\label{sup-tbl:participants}
\vspace{-0.1in}
\end{table}

%% file: tables/sup-dataset.tex
\begin{table*}[htbp]
\centering 
\resizebox{\linewidth}{!}{%
\begin{tabular}{|c|c c c c c c c c c|}
  \hline 
  \textbf{Dataset} & \textbf{Videos} & \textbf{Accidents} & \textbf{Events} & \textbf{Gaze providers} & \textbf{Duration(hrs)}  & \textbf{Number of Frames} & \textbf{Annotation type} &  \textbf{Gaze pattern~(per frame)} & \textbf{Fixations}\\
  \hline
  \hline
  EyeCar & 21 & 21 & rear-end collisions & 20 & 3.5  & 315K & spatial and temporal &  raw and average & 1,823,159 \\
 \hline 
\end{tabular}
}
\caption{The EyeCar dataset detailed information.}
\label{sup-tbl:eyecar}
\vspace{-0.1in}
\end{table*}

%% file: tables/sup-videos.tex
\begin{table*}[t!]

\centering 
\small\addtolength{\tabcolsep}{-2pt}
    \scalebox{0.9}{
    \begin{tabular}{|c| c c c c|}
      \hline 
      \textbf{Data Source} & \textbf{Feature} & \textbf{Type} & \textbf{Values} & \textbf{Scale}\\
      \hline
      \hline
      \multirow{2}{*}{videos}
      & day light & categorical & 0/1 & per frame\\
      & speed & categorical & slow, normal, fast & per frame\\
      \hline
      \multirow{2}{*}{eye's information} 
      & distance & categorical & near-reach, medium-reach, far-reach & per fixation in a frame\\ 
      & fixation duration & integer & [110ms-447ms] & per fixation in a frame\\
     \hline 
    \end{tabular}}
    \vspace{-0.1in}
    \caption{Detailed information about the videos and the fixations on each frame of each video.}
    \label{sup-tbl:videos}
\end{table*}

%% file: tables/ablation.tex
\begin{table}[t!]
\centering
\resizebox{\columnwidth}{!}{
\begin{tabular}{|l||c|c|c|c|c|c|c|c|c|}
\hline
\multirow{2}{*}{\diagbox{Ablated versions}{Dataset}} & \multicolumn{3}{c|}{EyeCar} & \multicolumn{3}{c|}{BDD-A}\\\cline{2-7}
&   CC $\uparrow$  &   KLD $\downarrow$ & $F_{\beta}\uparrow$  & CC $\uparrow$   & KLD $\downarrow$  &   $F_{\beta} \uparrow$ \\\hline \hline
-general features & 0.36 & 3.55 & 0.21 & 0.41 & 3.51 & 0.27 \\\hline 
-driving-related features & 0.69 & 2.21 & 0.30 & 0.60 & 2.07 & 0.39 \\\hline
\textbf{MEDIRL}  & \textbf{0.84} & \textbf{0.81} & \textbf{0.61} & \textbf{0.89} & \textbf{0.88} & \textbf{0.78}\\\hline
\end{tabular}
}\vspace{-0.1in}
\caption{Ablative study of MEDIRL using different combination of modules. The model used here is trained on BDD-A dataset and tested on EyeCar and BDD-A test dataset.}
\label{tbl:ablation}
\end{table}
\vspace{-0.1in}


 
